\documentclass[10pt,journal,compsoc]{IEEEtran}
\ifCLASSOPTIONcompsoc
  \usepackage[nocompress]{cite}
\else
  \usepackage{cite}
\fi
\ifCLASSINFOpdf
\else
\fi

\usepackage{graphicx}
\usepackage{amsmath,amssymb} %
\usepackage[ruled,vlined]{algorithm2e}
\usepackage{url}
\usepackage{epstopdf}
\usepackage{multirow}

\newcommand{\SEGNET}{temporal segment network}
\graphicspath{{figures/}}

\begin{document}
%
\title{Temporal Segment Networks for Action Recognition in Videos}
%
%
%
%

\author{Limin~Wang,
        Yuanjun~Xiong,
        Zhe~Wang,
        Yu~Qiao,
        Dahua~Lin,
        Xiaoou~Tang,
        and~Luc~Van~Gool
\IEEEcompsocitemizethanks{
\IEEEcompsocthanksitem Limin Wang and Luc Van Gool are with the Computer Vision Laboratory, ETH Zurich, Zurich, Switzerland. 
\IEEEcompsocthanksitem Yuanjun Xiong, Dahua Lin, and Xiaoou Tang are with the Department of Information Engineering, The Chinese University of Hong Kong, Hong Kong. 
\IEEEcompsocthanksitem Zhe Wang and Yu Qiao are with Shenzhen Institutes of Advanced Technology, Chinese Academy of Sciences, Shenzhen, China. 
\IEEEcompsocthanksitem Limin Wang and Yuanjun Xiong contribute equally to this paper.

}
}

\IEEEtitleabstractindextext{%
\begin{abstract}
  Deep convolutional networks have achieved great success for image recognition. However, for action recognition in videos, their advantage over traditional methods is not so evident. We present a general and flexible video-level framework for learning action models in videos. This method, called \SEGNET~(TSN), aims to model long-range temporal structures with a new segment-based sampling and aggregation module. This unique design enables our TSN to efficiently learn action models by using the whole action videos. The learned models could be easily adapted for action recognition in both trimmed and untrimmed videos with simple average pooling and multi-scale temporal window integration, respectively. We also study a series of good practices for the instantiation of \SEGNET~framework given limited training samples. Our approach obtains the state-the-of-art performance on four challenging action recognition benchmarks: HMDB51 ($71.0\%$), UCF101 ($94.9\%$), THUMOS14 ($80.1\%$), and ActivityNet v1.2 ($89.6\%$). Using the proposed RGB difference for motion models, our method can still achieve competitive accuracy on UCF101 ($ 91.0 \% $) while running at $ 340 $ FPS. Furthermore, based on the temporal segment networks, we won the video classification track at the ActivityNet challenge 2016 among 24 teams, which demonstrates the effectiveness of \SEGNET~and the proposed good practices. 
\end{abstract}

}

\maketitle

\IEEEdisplaynontitleabstractindextext

%
\IEEEpeerreviewmaketitle

\IEEEraisesectionheading{\section{Introduction}\label{sec:intro}}
Video-based action recognition has drawn considerable attention from the academic community~\cite{SimonyanZ14,WangS13a,WangQT13a,Ng15,WangQT15a}, owing to its applications in many areas like security and behavior analysis. For action recognition in videos, there are two crucial and complementary cues: appearances and temporal dynamics. The performance of a recognition system depends, to a large extent, on whether it is able to extract and utilize relevant information therefrom.
However, extracting such information is non-trivial due to a number of difficulties, such as scale variations, view point changes, and camera motions.
Thus it becomes crucial to design effective representations to tackle these challenges while learning categorical information of action classes.

Recently, Convolutional Neural Networks (ConvNets)~\cite{lecun-98} have achieved great success in classifying images of objects~\cite{KrizhevskySH12,SimonyanZ14a,SzegedyLJSRAEVR14}, scenes~\cite{ZhouLXTO14,ShenLH16,WangGHX016}, and complex events~\cite{XiongY2015,WangW0G16}.
ConvNets have also been introduced to solve the problem of video-based action recognition~\cite{KarpathyTSLSF14,SimonyanZ14,TranBFTP15,ZhangWWQW16}.
Deep ConvNets come with excellent modeling capacity and are capable of learning discriminative representations from raw visual data in large-scale supervised datasets (e.g., ImageNet~\cite{DengDSLL009}, Places~\cite{ZhouLXTO14}). However, unlike image classification, improvement brought by end-to-end deep ConvNets remains limited compared with traditional hand-crafted features for video-based action recognition. 

We argue that the application of ConvNets to action recognition in unconstrained videos is impeded by three major obstacles.
First, although long-range temporal structure has been proven crucial for understanding the dynamics in traditional methods \cite{NieblesCF10,GaidonHS13,WangQT14a,FernandoGMGT15}, it has not been considered as a critical factor in deep ConvNet frameworks~\cite{KarpathyTSLSF14,SimonyanZ14,TranBFTP15}. These methods usually focus on appearances and short-term motions (i.e., up to 10 frames), thus lacking the capacity to incorporate long-range temporal structure.
Recently there are a few attempts~\cite{varol,Ng15,DonahueJ2015} to deal with this problem. These methods mostly rely on dense temporal sampling with a pre-defined sampling interval, which would incur excessive computational cost when applied to long videos. More importantly, the limited memory space available severely limits the duration of video to be modeled. This poses a risk of missing important information for videos longer than the affordable sampling duration.

Second, existing action recognition methods were mostly devised for trimmed videos. However, to deploy the learned action models in a realistic setting we often need to deal with untrimmed videos (e.g., THUMOS~\cite{THUMOS}, ActivityNet~\cite{HeilbronEGN15}), where each action instance may only occupy a small portion of the whole video. The dominating background portions may interfere with the prediction of action recognition models. To mitigate this issue, we need to take account of localizing action instances and avoiding the influence of background video at the same time. Therefore, it is a non-trivial task to apply the learned action models to action recognition in untrimmed videos.

Third, training action recognition models often meets a number of practical difficulties: 1) training deep ConvNets usually requires a large volume of training samples to achieve optimal performance. However, publicly available action recognition datasets (e.g., UCF101~\cite{Soomro12}, HMDB51~\cite{KuehneJGPS11}) remain limited in both size and diversity, making the model training prone to over-fitting. 2) optical flow extraction to capture short-term motion information becomes a computational bottleneck for deploying the learned models to large-scale action recognition datasets.

These challenges motivate us to study the action recognition problem in this paper from the following three aspects : \emph{1) how to effectively learn video representation that captures long-range temporal structure; 2) how to exploit these learned ConvNet models for the more realistic setting of untrimmed videos; 3) how to efficiently learn the ConvNet models given limited training samples and apply them on large scale data.} 

To capture long-range temporal structure, we develop a modular video-level architecture, called \emph{\SEGNET}~(TSN), which provides a conceptually simple, flexible, and general framework for learning action models in videos. It is based on our observation that {\em consecutive frames are highly redundant, where a sparse and global temporal sampling strategy would be more favorable and efficient in this case.} The TSN framework first extracts short snippets over a long video sequence with a sparse sampling scheme, where the video is first divided into a fixed number of segments and one snippet is randomly sampled from each segment.
Then, a segmental consensus function is employed to aggregate information from the sampled snippets. By this means, \SEGNET s can model long-range temporal structures over the whole video, in a way that its computational cost is independent of the video duration. In practice, we comprehensively study the effect of different segment numbers and propose five aggregation functions to summarize the prediction scores from these sampled snippets, including three basic forms: average pooling, max pooling, and weighted average, as well as two advanced schemes: top-$\mathcal{K}$ pooling and adaptive attention weighting. The latter two are designed to automatically highlight discriminative snippets while reducing the impact of less relevant ones during training, thus contribute to a better learned action model.

To apply the action models learned by TSN to untrimmed videos, we design a hierarchical aggregating strategy, called Multi-scale Temporal Window Integration (M-TWI), to yield the final prediction results for untrimmed videos. Most of previous action recognition methods are constrained to classify manually trimmed video clips. However, this setting may be impractical and unrealistic, as videos on the web are untrimmed by nature and manually trimming these videos is labor demanding. Following the idea of \SEGNET~framework, we first divide the untrimmed video into a sequence of short windows of fixed duration. We then perform action recognition for each window independently by max pooling over these snippet-level recognition scores inside this window. Finally, following the aggregation function of \SEGNET~framework, we employ the top-$\mathcal{K}$ pooling or attention weighting to aggregate the predictions from these windows to produce the video-level recognition results.
Due to its capability of implicitly selecting intervals with discriminative action instances while suppressing the influence of noisy background, this newly designed aggregation module is effective for untrimmed video recognition

To tackle the practical difficulties in learning and applying action recognition models, we discover a number of good practices to resolve the issues caused by the limited training samples, and perform a systematical study over the input modalities to unleash the full potential of ConvNets for action recognition. Specifically, we first propose a {\em cross-modality initialization} strategy to transfer the learned representations from RGB modality to other modalities like optical flow and RGB difference. Second, we develop a principled method to perform Batch Normalization (BN) in a fine-tuning scenario, denoted as {\em partial BN}, where only the mean and variance of first BN layer are updated adaptively to handle domain shift. Moreover, to fully utilize visual content from videos, we empirically study four types of input modalities with our \SEGNET~framework, namely a single RGB image, stacked RGB difference, stacked optical flow field, and stacked warped optical flow field. Combining RGB and RGB difference, we build the best-ever real-time action recognition system, which has numerous potential applications in real-world problems. 

We perform experiments on four challenging action recognition datasets, namely HMDB51~\cite{KuehneJGPS11}, UCF101~\cite{Soomro12}, THUMOS~\cite{THUMOS}, and ActivityNet~\cite{HeilbronEGN15}, to verify the effectiveness of our method for action recognition in both trimmed and untrimmed videos. In experiments, models learned using the temporal segment network significantly outperform the state of the art on these four challenging action recognition benchmark datasets. Additionally, following the basic \SEGNET~framework, we further improve our action recognition method by introducing the latest deep model architectures (e.g., ResNet~\cite{HeZRS15} and Inception V3~\cite{SzegedyVISW16}), and incorporating the audio as a complementary channel. Our final action recognition method secures the 1st place in untrimmed video classification at the ActivityNet Large Scale Activity Recognition Challenge 2016. We also visualize the our learned two-stream models trying to provide insights into how they work. These visualized models also justify the effectiveness of our \SEGNET~framework qualitatively.

Overall, we analyze different aspects of the problems in efficiently and effectively learning and applying action recognition models and make \textbf{three major contributions}: 1) we propose an end-to-end framework, dubbed temporal segment network (TSN), for learning video representation that captures long-term temporal information; 2) we design a hierarchical aggregation scheme to apply action recognition models to untrimmed videos; 3) we investigate a series of good practices for learning and applying deep action recognition models. 

This journal paper extends our previous work~\cite{WangXWQLTV16} in a number of aspects. {\em First}, we introduce new aggregation functions into the \SEGNET~framework, which turn out to be effective to highlight important snippets while suppress background noise. {\em Second}, we extend the original action recognition pipeline to untrimmed video classification, by designing a hierarchical aggregating strategy. {\em Third}, we add more exploration studies on the different aspects of  \SEGNET~framework and more experimental investigation on two new datasets (i.e., THUMOS15 and ActivityNet). {\em Finally}, based on our \SEGNET~framework, we present an  effective and efficient action recognition solution for ActivtyNet Large Scale Activity Challenge 2016, which ranks \#1 in untrimmed video classification among 24 teams, and give a detailed analysis on different components of our method to highlight the important ingredients. The code of our method and learned models are publicly available to facilitate future research \footnote{\url{https://github.com/yjxiong/temporal-segment-networks/}}.

\section{Related Work}
\label{sec:rw}
Action recognition has been studied extensively in recent years and readers can refer to \cite{ForsythAIOR05,TuragaCSU08,AggarwalR11} for good surveys. Here, we only cover the work related to our methods.

\subsection{Video Representation}

 For action recognition in videos, the visual representation plays a crucial role. We roughly categorize the related action recognition approaches into two types: methods based {\em hand-crafted} features and those using {\em deeply-learned} features.

{\bf Hand-crafted features}. In recent years, researchers have developed many different spatio-temporal feature detectors for video, such as 3D-Harris~\cite{Laptev05}, 3D-Hessian~\cite{WillemsTG08}, Cuboids~\cite{Dollar05}, Dense Trajectories~\cite{WangKSL11}, Improved Trajectories~\cite{WangS13a}. Usually, a local 3D-region is extracted around the interest points or trajectories, and a histogram descriptor is computed to capture the appearance and motion information, such as Histogram of Gradient and Histogram of Flow (HOG/HOF)~\cite{LaptevMSR08}, Histogram of Motion Boundary (MBH) \cite{WangKSL11}, 3D Histogram of Gradient (HOG3D)~\cite{KlaserMS08}, Extended SURF (ESURF)~\cite{WillemsTG08}, and so on. Then encoding methods are employed to aggregate these local descriptors into a global representation, and typical encoding methods include Bag of Visual Words (BoVW)~\cite{csurka2004}, Fisher vector (FV)~\cite{SanchezPMV13}, Vector of Locally Aggregated Descriptors (VLAD)~\cite{JegouPDSPS12}, and Multi-View Super Vector (MVSV)~\cite{CaiWPQ14}. These local features share the merits of locality and simplicity, but may lack semantic and discriminative capacity.

To overcome the limitation of local descriptors, several mid-level representations have been proposed for action recognition \cite{RaptisKS12,JainGRD13,WangQT13a,ZhangZD13,ZhuWYZT13,WangQT15b,SadanandC12}. Raptis \emph{et al.} \cite{RaptisKS12} grouped similar trajectories into clusters, each of which was regarded as an action part.  Jain \emph{et al.} \cite{JainGRD13} extended the idea of discriminative patches into videos and proposed discriminative spatio-temporal patches for representing videos. Zhang \emph{et al.} \cite{ZhangZD13} proposed to discover a set of mid-level patches in a strongly-supervised manner. Similar to 2-D poselet \cite{BourdevM09}, they tightly clustered action parts using human joint labeling, dubbed \emph{acteme}.  Wang \emph{et al.}~\cite{WangQT13a} proposed a data-driven approach to discover those effective parts with high motion salience, known as {\em motionlet}. Zhu \emph{et al.} \cite{ZhuWYZT13} proposed a two-layer \emph{acton} representation for action recognition. The weakly-supervised actons were learned via a max-margin multi-channel multiple instance learning framework. Wang \emph{et al.}~\cite{WangQT15b} proposed a multiple level representation called as {\em MoFAP} by concatenating motion features, atoms, and phrases. Sadanand {\em et al.}~\cite{SadanandC12} presented a high-level video representation called as {\em Action Bank} by using a set action templates to describe the video content. In summary, these mid-level representations have the merits of representative and discriminative power, but still depends on the low-level hand-crafted features.

{\bf Deeply-learned features}. Several works have been trying to learn deep features and design effective ConvNet architectures for action recognition in videos~\cite{KarpathyTSLSF14,SimonyanZ14,TranBFTP15,JiXYY13,SunJYS15,WangQT15a,Ng15,DonahueJ2015,WuWJYX15,varol}. Karpathy \emph{et al.}~\cite{KarpathyTSLSF14} first tested ConvNets with deep structures on a large dataset (Sports-1M). Simonyan \emph{et al.}~\cite{SimonyanZ14} designed two-stream ConvNets containing spatial and temporal nets by exploiting ImageNet dataset for pre-training and calculating optical flow to explicitly capture motion information. Tran \emph{et al.}~\cite{TranBFTP15} explored 3D ConvNets~\cite{JiXYY13} on the realistic and large-scale video datasets, where they tried to learn spatio-temporal features with the operations of 3D convolution and pooling. Sun \emph{et al.}~\cite{SunJYS15} proposed a factorized spatio-temporal ConvNets and exploited different ways to decompose 3D convolutional kernels. Wang {\em et al.}~\cite{WangQT15a} proposed a hybrid representation by using trajectory-pooled deep-convolutional descriptors (TDD), which share the merits of improved trajectories~\cite{WangS13a} and two-stream ConvNets~\cite{SimonyanZ14}.  Feichtenhofer \emph{et al.}~\cite{FeichtenhoferPZ16} further extended the two-stream ConvNets with convolutional fusion of two streams. Several works~\cite{Ng15,WuWJYX15,DonahueJ2015} tried to use recurrent neural networks (RNN), in particular LSTM, to model the temporal evolution of frame features for action recognition in videos.

Our work is related to those deep learning methods. In fact, any existing ConvNet architecture can work with TSN framework, and thus be combined with the proposed sparse sampling strategy and aggregation functions to enhance the modeling capacity with long-range information. Meanwhile, our \SEGNET~is an end-to-end architecture, where the model parameters could be jointly optimize with the standard back propagation algorithm.

\subsection{Temporal Structure Modeling}

Many research works have been devoted to modeling the temporal structure of video for action recognition~\cite{NieblesCF10,GaidonHS13,WangQT14a,PirsiavashR14,WangQT14b,FernandoGMGT15}. Gaidon \emph{et al.}~\cite{GaidonHS13} annotated each atomic action for each video and proposed Actom Sequence Model (ASM) for action detection. Niebles \emph{et al.}~\cite{NieblesCF10} proposed to use latent variables to model the temporal decomposition of complex actions, and resorted to the Latent SVM~\cite{FelzenszwalbGMR10} to learn the model parameters in an iterative approach. Wang \emph{et al.}~\cite{WangQT14a} and Pirsiavash \emph{et al.}~\cite{PirsiavashR14} extended the temporal decomposition of complex action into a hierarchical manner using Latent Hierarchical Model (LHM) and Segmental Grammar Model (SGM), respectively. Wang \emph{et al.}~\cite{WangQT14b} designed a sequential skeleton model (SSM) to capture the relations among dynamic-poselets, and performed spatio-temporal action detection. Fernando~\cite{FernandoGMGT15} modeled the temporal evolution of BoVW representations for action recognition. 

Several recent works focused on modeling long-range temporal structure with ConvNets~\cite{Ng15,varol,DonahueJ2015}. In general, these methods directly operated on a continuous video frame sequence with recurrent neural networks~\cite{Ng15,DonahueJ2015,WuWJYX15} or 3D ConvNets~\cite{varol}. Although these methods aim to deal with longer video duration, they usually process sequences of fixed lengths ranging from 64 to 120 frames due to the limit of computational cost and GPU memory. It is still non-trivial for these methods to learn from the entire video due to their limited temporal coverage. Our method differs from these end-to-end deep ConvNets by its novel adoption of a sparse temporal sampling strategy, which enables efficient learning using the entire videos without the limitation of sequence length. Therefore, our \SEGNET~is a video-level and end-to-end framework for temporal structure modeling on the entire video.

\section{Temporal Segment Networks}
\label{sec:vds}

In this section, we give a detailed description of our \SEGNET~framework. Specifically, we first discuss the motivation of segment based sampling. Then, we introduce the architecture of \SEGNET~framework. After this, we present several aggregating functions of \SEGNET~and provide analysis on these functions. Finally, we investigate several practical issues for the instantiation of \SEGNET~framework.

\subsection{Motivation of Segment Based Sampling}

As discussed in Sec.~\ref{sec:intro}, long-range temporal modeling is important for action understanding in videos. The existing deep architectures such as two-stream ConvNets~\cite{SimonyanZ14} and 3D convolutional networks~\cite{TranBFTP15} are designed to operate on a single frame or a stack of frames (e.g., 16 frames) with limited temporal durations. Therefore, these structures lack capacity of incorporating long-range temporal information of videos into the learning of action models.

In order to model long-range temporal structures, several approaches have been proposed, such as stacking more consecutive frames (e.g., 64 frames~\cite{varol}) or sampling more frames at a fixed rate (e.g., 1FPS~\cite{Ng15}). Although this {\em dense} and {\em local} sampling could help to relieve the problem of the original short-term CovNets~\cite{SimonyanZ14,TranBFTP15}, it still suffers in both {\em computational} and {\em modeling} aspects. From the computational perspective, it would greatly increase the cost of ConvNet training, as this dense sampling usually requires a large number of frames to capture long-range structures. For example, it totally samples 64 frames in the work of~\cite{varol} and 120 frames for the method of~\cite{Ng15}. From the modeling perspective, its temporal coverage is still local and limited by its fixed sampling interval, failing to capture the visual content over the entire video. For instance, the sampled 64 frames~\cite{varol} only occupy a small portion of a 10-second video (around 300 frames).

\begin{figure*}[t]
  \centering
  \includegraphics[width=.98\textwidth]{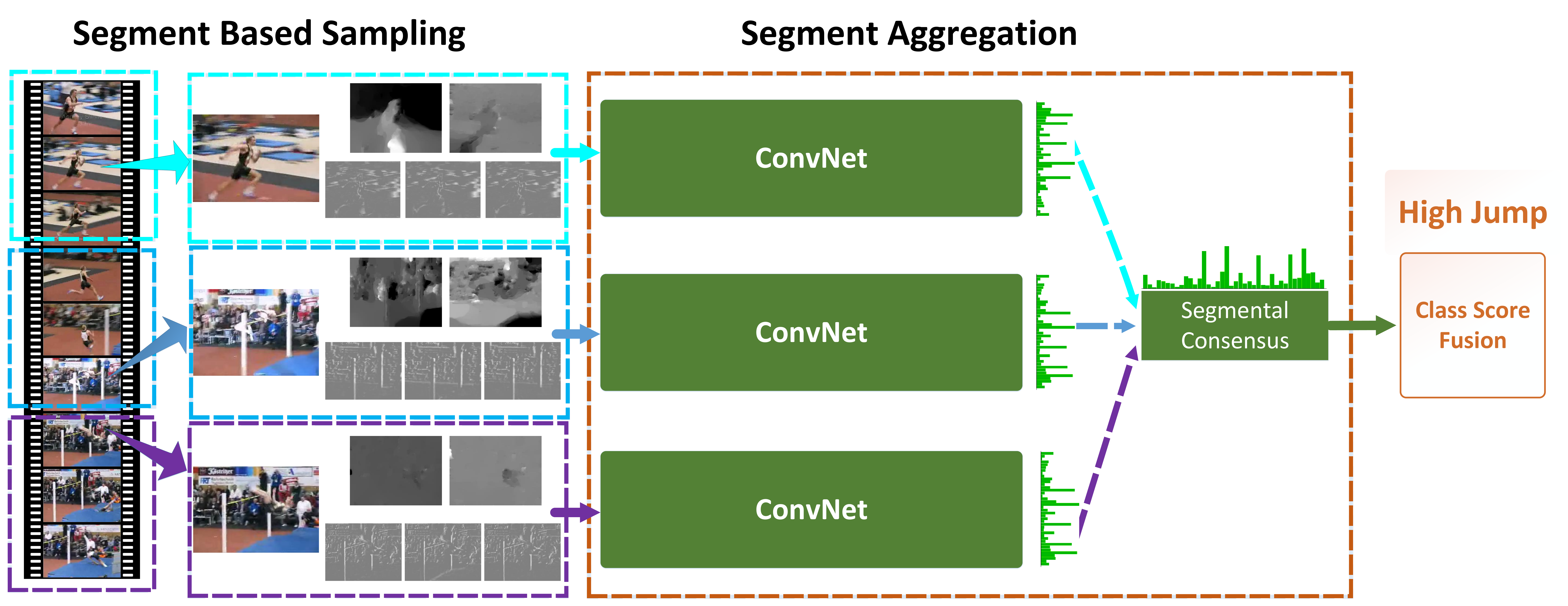}
  \vspace{-10pt}
  \caption{Temporal segment network: One input video is divided into $ K $ segments (here we show the $ K=3 $ case) and a short snippet is randomly selected from each segment. The snippets are represented by modalities such as RGB frames, optical flow (upper grayscale images), and RGB differences (lower grayscale images). The class scores of different snippets are fused by an the segmental consensus function to yield segmental consensus, which is a video-level prediction. Predictions from all modalities are fused to produce the final prediction. ConvNets on all snippets share parameters.}
  \label{fig:pipeline}
  \vspace{-15pt}
\end{figure*}

We observe that {\em although the frames are densely recorded in the videos, the content changes relatively slowly}. This motivates us to explore a new paradigm for temporal structure modeling, called \emph{segment based sampling}. This strategy is essentially a kind of {\em sparse} and {\em global} sampling. 
Concerning the property of spareness, only a small number of sparsely sampled snippets would be used to model the temporal structures in a human action. Normally, the number of sampled frames for one training iteration is fixed to a pre-defined value independent of the durations of the videos. This guarantees that the computational cost will be constant, regardless of the temporal range we are dealing with.
On the global property, our segment based sampling ensures these sampled snippets would distribute uniformly along the temporal dimension. Therefore, no matter how long the action videos will last for, our sampled snippets would always roughly cover the visual content of whole video, enabling us to model the long-range temporal structure throughout the entire video.
Based on this paradigm for temporal structure modeling, we propose \SEGNET, a video-level framework as shown in Figure \ref{fig:pipeline}, which would be explained in the next subsection.

\subsection{Framework and Formulation}\label{sec:seg}

We aim to design an effective and efficient video-level framework, coined \emph{Temporal Segment Network} (TSN), by using a new strategy of segment based sampling. Instead of working on a single frame or a short frame stack, \SEGNET s operate on a sequence of short snippets sampled from the entire video. 
To make these sampled snippets represent the contents of the whole video while still keeping reasonable computational cost, our segment based sampling first divides the video into several segments of equation duration, and then one snippet is randomly sampled from its corresponding segment.
Each snippet in this sequence produces its own snippet-level prediction of the action classes, and an consensus function is designed to aggregate these snippet-level predictions into the video-level scores. This video-level score is more reliable and informative than the original snippet-level prediction, since it captures the long-range information over the entire video.
During the training process, the optimization objectives are defined on the video-level predictions and optimized by iteratively updating the model parameters.

Formally, given a video $V$, we divide it into $K$ segments $\{S_1, S_2, \cdots, S_K\}$ of equal durations. One snippet $ T_k $ is randomly sampled from its corresponding segment $S_k$. Then, the temporal segment network models a sequence of snippets $(T_1, T_2, \cdots, T_K)$  as follows:
\begin{equation}
\begin{split}
& \mathrm{TSN}(T_1, T_2, \cdots, T_K) = \\
& \mathcal{H} (\mathcal{G}(\mathcal{F}(T_1;\mathbf{W}), \mathcal{F}(T_2;\mathbf{W}), \cdots, \mathcal{F}(T_K; \mathbf{W}))).
\label{equ:model}
\end{split}
\end{equation}
Here, $\mathcal{F}(T_k; \mathbf{W})$ is the function representing a ConvNet with parameters $\mathbf{W}$ which operates on the short snippet $T_k$ and produces class scores over all the classes. The segmental consensus function $\mathcal{G}$ combines the outputs from multiple short snippets to obtain a consensus of class hypothesis among them. Based on this consensus, the prediction function $\mathcal{H}$ predicts the probability of each action class for the whole video. Here we choose the widely used Softmax function for $\mathcal{H}$.  In our \SEGNET~framework, the form of consensus function $ \mathcal{G} $ is of great importance, as it should be equipped with high modeling capacity while still could be differentiable or at least has subgradients. The high modeling capacity refers to the ability to effectively aggregate snippet-level prediction into video-level scores and the differentiability allows our \SEGNET~framework to be easily optimized using backpropagation. We will provide the details on these consensus functions in the next subsection.

Combining standard categorical cross-entropy loss, the final loss function regarding the segmental consensus $ \mathbf{G} = \mathcal{G}(\mathcal{F}(T_1;\mathbf{W}), \mathcal{F}(T_2;\mathbf{W}), \cdots, \mathcal{F}(T_K; \mathbf{W})) $ is formed as
\begin{equation}
\mathcal{L}(y, \mathbf{G}) = -\sum_{i=1}^{C}y_i \left(g_i - \log\sum_{j=1}^{C}\exp{g_j}\right),
\end{equation}
where $ C $ is the number of action classes, $ y_i $ the groundtruth label concerning class $ i $, and $g_j$ the $j^{th}$ dimension of $\mathbf{G}$. During the training phase of our \SEGNET~framework, the gradients of the loss value $\mathcal{L}$ with respect to model parameters $ \mathbf{W} $  can be derived as
\begin{equation}\label{eq:grad}
\frac{\partial \mathcal{L}(y, \mathbf{G})}{\partial \mathbf{W}} = 
\frac{\partial\mathcal{L}}{\partial \mathbf{G}} \sum_{k=1}^K \frac{\partial \mathbf{G}}{\partial \mathcal{F}(T_k)} \frac{\partial \mathcal{F}(T_k)}{\partial \mathbf{W}},
\end{equation}
where $ K $ is number of segments in \SEGNET. When we use a gradient-based optimization method, such as stochastic gradient descent (SGD), to learn the model parameters, Eq.~\ref{eq:grad} shows that the parameter updates are utilizing the segmental consensus $ \mathbf{G} $ derived from all snippet-level predictions.
In this sense, \SEGNET~can learn model parameters from the entire video rather than a short snippet. 
Furthermore, by fixing $ K $ for all videos, we assemble a sparse temporal sampling to select a small number of snippets.
It drastically reduces the computational cost for evaluating ConvNets on the frames, compared with previous works using densely sampled frames~\cite{Ng15,varol,DonahueJ2015}.

\subsection{Aggregation Function and Analysis}

As analyzed above, the consensus (aggregation) function is an important component in our \SEGNET~framework. In this subsection, we give a detailed description about the design of aggregation functions and derive their gradients with respect to snippet-level prediction scores. We also analyze the properties of different kinds of aggregation functions and provide some modeling insight. Specifically, we propose five types of aggregation functions: max pooling, average pooling, top-$\mathcal{K}$ pooling,  weighted average, and attention weighting.

{\bf Max pooling.} In this aggregation function, we apply max pooling to the prediction score of each category among the sampled snippets, i.e., $g_i = \max_{k \in \{1,2,\cdots, K\}} f_i^k$, where $f_i^k$ is the $i^{th}$ element of $\mathbf{F}^k$ = $\mathcal{F}(T_k; \mathbf{W})$. The gradient of $g_i$ with respect to $f_i^k$ can be easily computed as:
\begin{equation}
\frac{\partial g_i}{\partial f_i^k} =  \left \{ 
 \begin{array}{l}
 1, \ \textrm{if} \ \ k=\textrm{arg}\max_{l} f_i^l, \\
 0,\ \ \textrm{otherwise.}
 \end{array} 
 \right.
\end{equation}
The basic idea of max pooling is to seek a {\em single} and most discriminative snippet for each action class and utilize this strongest activation as the video-level response of this category. Intuitively, it devotes its emphasis to a single snippet, while completely ignoring the responses of other snippets. Thus, this aggregating function encourages \SEGNET~to learn from a most discriminative snippet for each action class, but lacks the capacity of jointly modeling multiple snippets for a video-level action understanding.

{\bf Average pooling.} One alternative to max pooling aggregation function is the average pooling, where we perform average operation over these snippet-level prediction scores for each class, i.e., $g_i = \frac{1}{K} \sum_{k=1}^K f_i^k$. The gradient of average aggregation function $g_i$ withe respect to $f_i^k$ is derived as follows:
\begin{equation}
\frac{\partial g_i}{\partial f_i^k} = \frac{1}{K}.
\end{equation}
The basic intuition behind average pooling is to leverage the responses of {\em all} snippets for action recognition, and use their mean activation as the video-level prediction. In this sense, average pooling is able to jointly model multiple snippets and capture the visual information from the whole video. On the other hand, in particular for noisy videos with complex background, some snippets may be action-irrelevant and averaging over these background snippets may hurt the final recognition performance.

{\bf Top-$\mathcal{K}$ pooling.} To strike a balance between max pooling and average pooling, we propose a new aggregation function, named {\em Top-$\mathcal{K}$ pooling}. In this aggregation function, we first select $\mathcal{K}$ most discriminative snippets for each action category and then perform average pooling over these selected snippets, i.e., $g_i = \frac{1}{\mathcal{K}} \sum_{k=1}^\mathcal{K} \alpha_k f_i^k$, where $\alpha_k$ is the indicator of selection, and is set as 1 if selected and otherwise 0. Max pooling and average pooling can be considered as special cases of top-$\mathcal{K}$ pooling, where $\mathcal{K}$ is set to $1$ or $K$, respectively. Similarly, the gradient of $g_i$ with respect to $f_i^k$ can be computed as follows:
\begin{equation}
\frac{\partial g_i}{\partial f_i^k} =  \left \{ 
 \begin{array}{l}
 \frac{1}{\mathcal{K}}, \ \textrm{if} \ \ \alpha_k = 1, \\
 0,\ \ \textrm{otherwise.}
 \end{array} 
 \right.
\end{equation}
Intuitively, this aggregation function is able to determine a subset of discriminative snippets adaptively for different videos. As a result, it shares merits of both max pooling and average pooling, having capacity of jointly modeling multiple relevant snippets while avoiding the influence of background snippets.

{\bf Linear weighting.} In this aggregation function, we aim to perform an element-wise weighted linear combination on the prediction score for each action category. Specifically, we define the aggregation function as $g_i = \sum_{k=1}^K \omega_k f_i^k$, where $w_k$ is the weight for the $k^{th}$ snippet. In this aggregation function, we introduce a model parameter $\omega$ and compute the gradients of $g_i$ with respect to $f_i^k$ and $w_k$ as follows:
\begin{equation}
\frac{\partial g_i}{\partial f_i^k} = \omega_k, \ \ \ \frac{\partial g_i}{\partial \omega_k} = f_i^k.
\end{equation}
In practice, we use this equation to update the network weights $\mathbf{W}$ and the combination weights $\omega$ alternatively. The basic assumption underlying this aggregation function is that action can be decomposed into several phases and these different phases may play different roles in recognizing action classes. This aggregation function is expected to learn importance weights of different phases of an action class. Compared with previous pooling based aggregation functions, this linear weighting acts as a soft version of snippet selection. 

{\bf Attention weighting.} It is obvious that the above linear weighting scheme is data independent, thus lacking the capacity of considering the difference between videos. To overcome this limitation, we propose an adaptive weighting method, called {\em attention weighting}. In this aggregation function, we aim to learn a function to automatically assign an importance weight to each snippet according to the video content. Formally, the aggregation function is defined as $g_i = \sum_{k=1}^K \mathcal{A}(T_k) f_i^k$, where $\mathcal{A}(T_k)$ is the attention weight for snippet $T_k$ and calculated according to video content adaptively. Within this formulation, we could calculate the gradient of $g_i$ with respect to $f_i^k$ and $\mathcal{A}(T_k)$ as follows:
\begin{equation}
\frac{\partial g_i}{\partial f_i^k} = \mathcal{A}(T_k), \ \ \ \frac{\partial g_i}{\partial \mathcal{A}(T_k)} = f_i^k.
\end{equation}

In this attention weighting scheme, the design of attention weighting function $\mathcal{A}(T_k)$ is crucial for final performance. In the current implementation, we first extract visual feature $\mathbf{R} = \mathcal{R}(T_k)$ from each snippet with the same ConvNet and then produce the attention weights as:
\begin{equation}
\begin{split}
e_k = \omega^{att} \mathcal{R}(T_k), \\
\mathcal{A}(T_k) = \frac{\exp(e_k)}{\sum_{l=1}^K \exp(e_l)},
\end{split}
\end{equation}
where $\omega^{att}$ is the parameter of attention weighting function and will be learned jointly with network weights $\mathbf{W}$. Here $\mathcal{R}(T_k)$ is the visual feature for the $k^{th}$ snippet. Currently it is the activation of last hidden layer. Within this formation, we can calculate the gradient of $\mathcal{A}(T_k)$ with respect to attention model parameter $\omega^{att}$ as:
\begin{equation}
\frac{\partial \mathcal{A}(T_k)} {\partial \omega^{att}} =  \sum_{l=1}^K \frac{\partial \mathcal{A}(T_k)}{\partial e_l} \mathcal{R}(T_l),
\end{equation}
where the gradient of $\frac{\partial \mathcal{A}(T_k)}{\partial e_l}$ is computed as:
\begin{equation}
 \frac{\partial \mathcal{A}(T_k)} {\partial e_l}= \left \{ 
 \begin{array}{l}
 \mathcal{A}(T_k)(1-\mathcal{A}(T_l)), \ \textrm{if} \ \ l=k, \\
 -\mathcal{A}(T_k)\mathcal{A}(T_l),\ \ \textrm{otherwise.}
 \end{array} 
 \right.
\end{equation}
Having this gradient formula, we can learn the attention model parameters $\omega^{att}$ using back-propagation together with the ConvNet parameters $\mathbf{W}$. In addition, due to the introduction of attention model $\mathcal{A}(T_k)$, the basic back-propagation formula in Eq.~\ref{eq:grad} should be rectified as follows:
\begin{equation}\label{eq:attgrad}
\frac{\partial \mathcal{L}(y, \mathbf{G})}{\partial \mathbf{W}} = 
\frac{\partial\mathcal{L}}{\partial \mathbf{G}} \sum_{k=1}^K \left ( \frac{\partial \mathbf{G}}{\partial \mathcal{F}(T_k)} \frac{\partial \mathcal{F}(T_k)}{\partial \mathbf{W}} + \frac{\partial \mathbf{G}}{\partial \mathcal{A}(T_k)} \frac{\partial \mathcal{A}(T_k)}{\partial \mathbf{W}} \right ).
\end{equation}
Overall, the advantages of introducing attention model $\mathcal{A}(T_k)$ come from two aspects: (1) The attention model enhances the modeling capacity of our \SEGNET~framework by automatically estimating the importance weight of each snippet based on the video content. (2) Due to the fact that the attention model is based on ConvNet representations $\mathbf{R}$, it leverages extra backpropagation information to guide the learning process of ConvNet parameter $\mathbf{W}$ and may accelerate the convergence of training.

\subsection{TSN in Practice}\label{sec:training}

Temporal segment network (TSN) provides a general framework to perform video-level learning. In order to train TSN models to achieve optimal performance, a few practical issues have to be taken into account.
To this end, we study a series of practical matters from the aspects of TSN architectures, TSN inputs, and TSN training.

\begin{figure}[t]
  \centering{
    \includegraphics[width=0.24\linewidth]{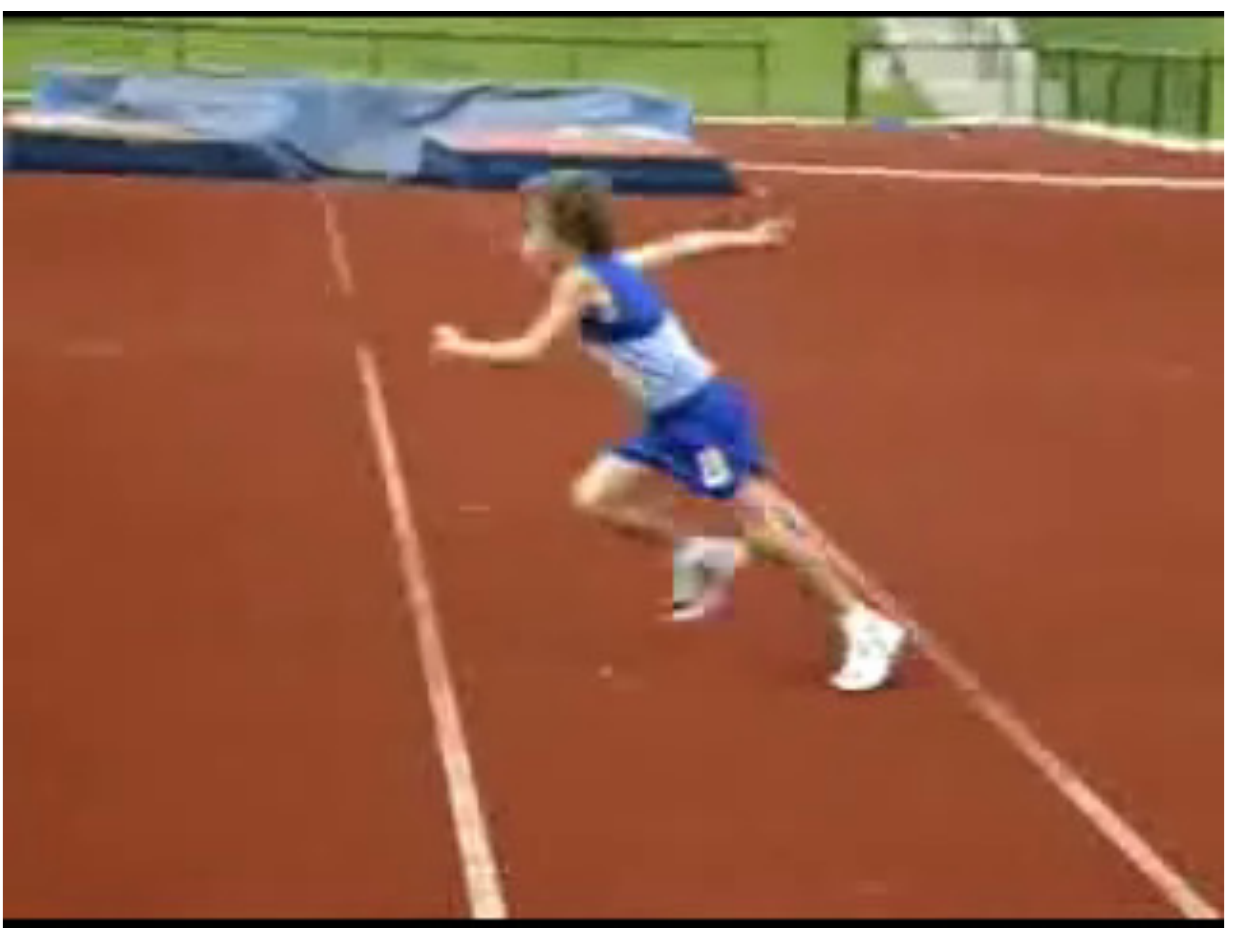}
    \includegraphics[width=0.24\linewidth]{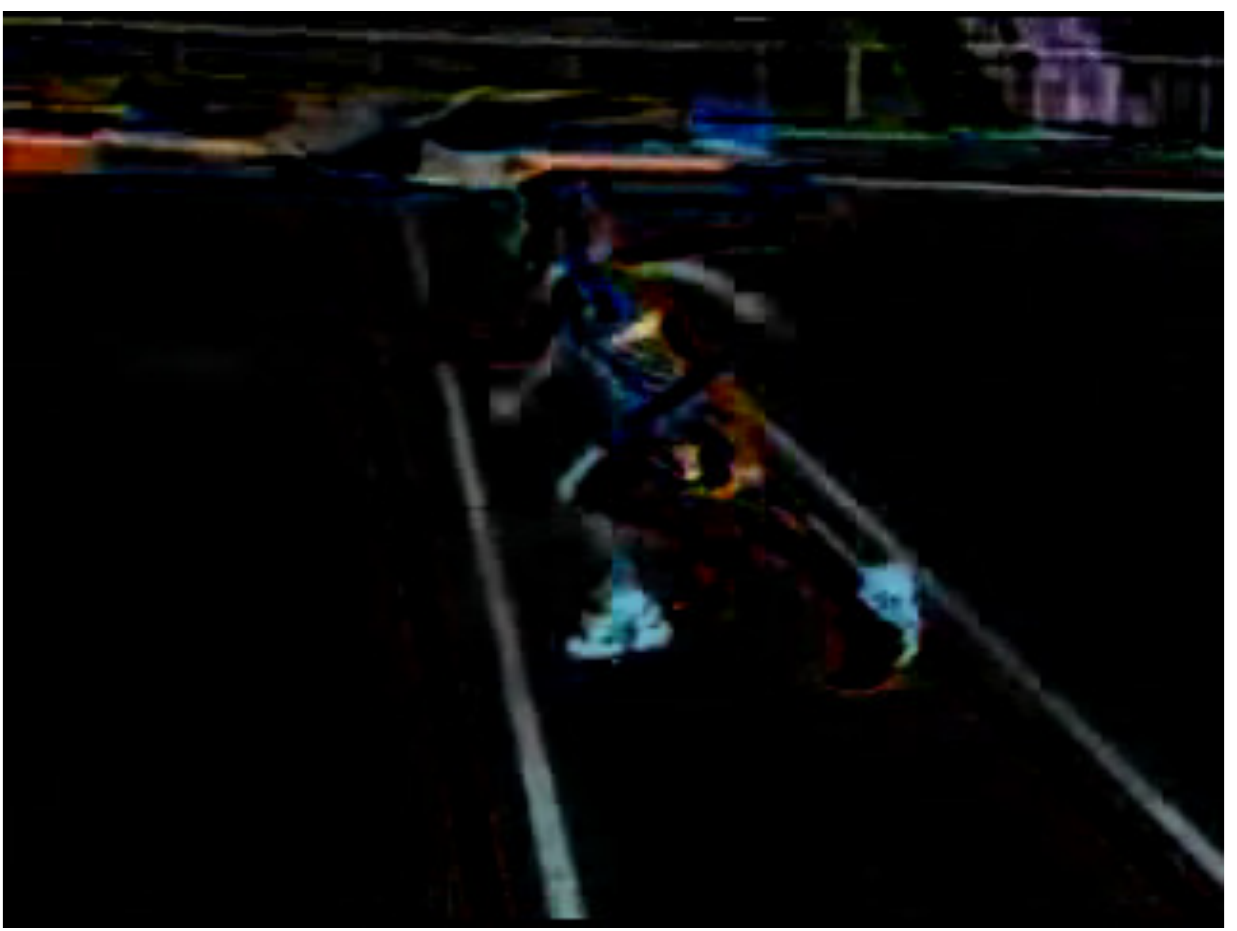}
    \includegraphics[width=0.24\linewidth]{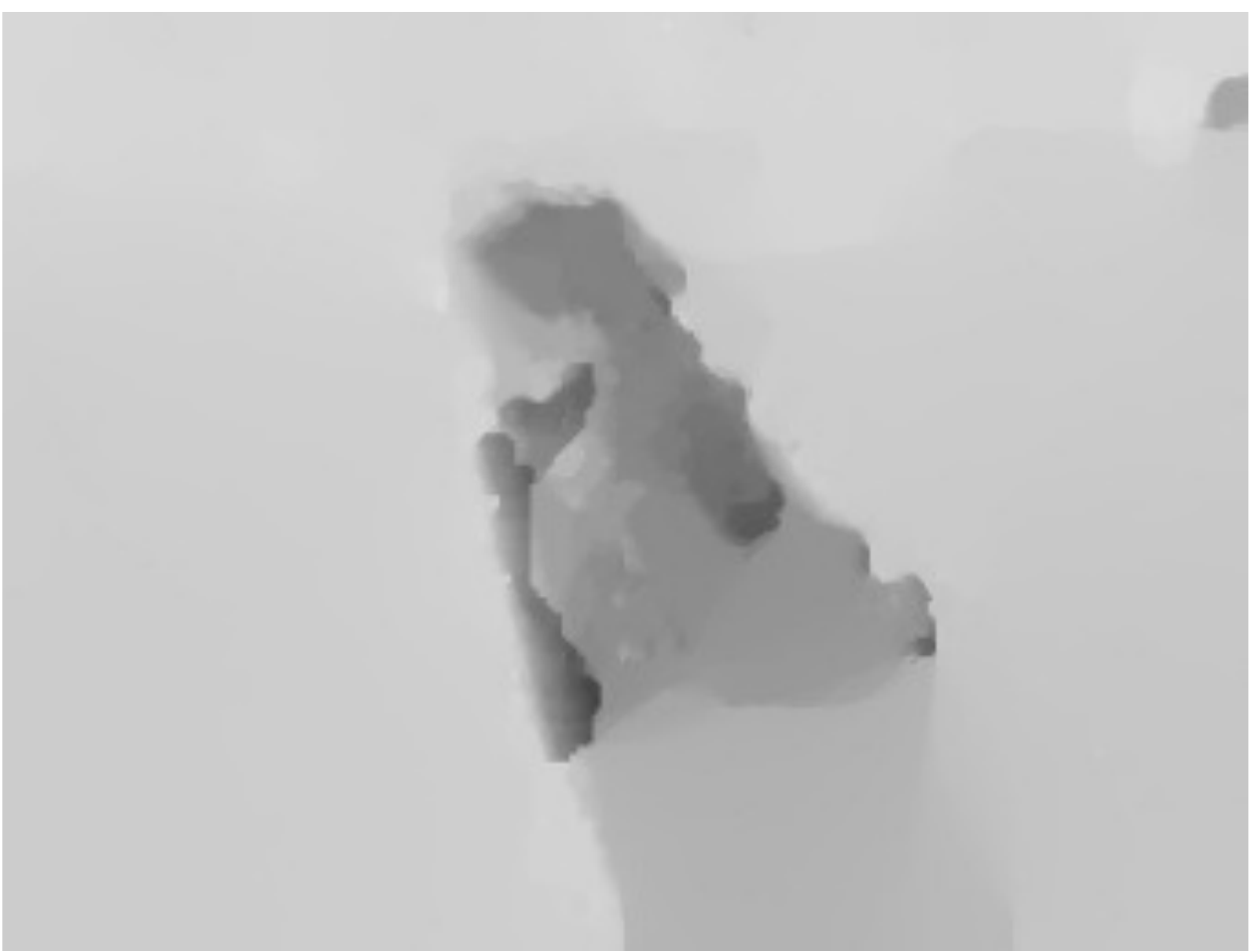}
    \includegraphics[width=0.24\linewidth]{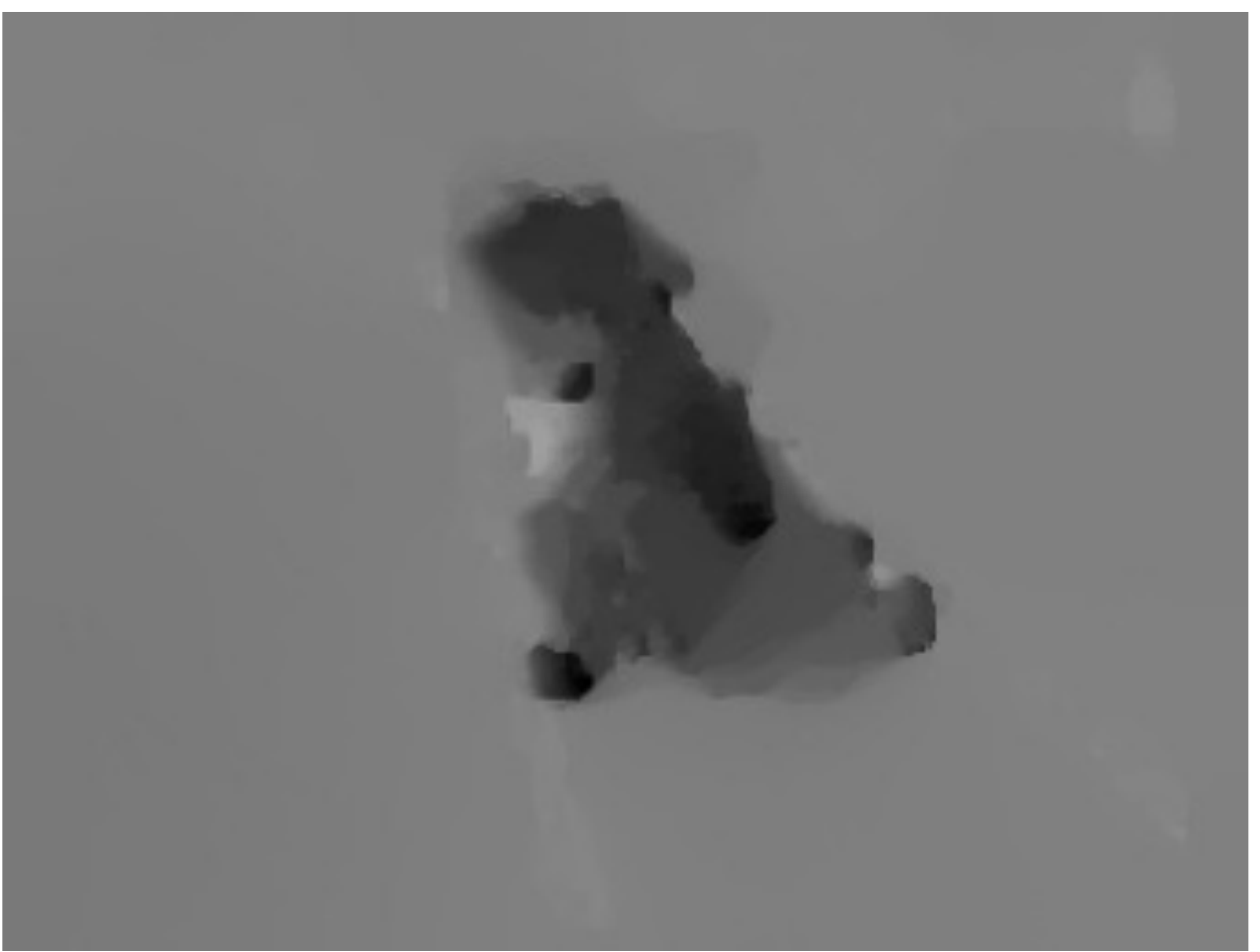} \\
    \includegraphics[width=0.24\linewidth]{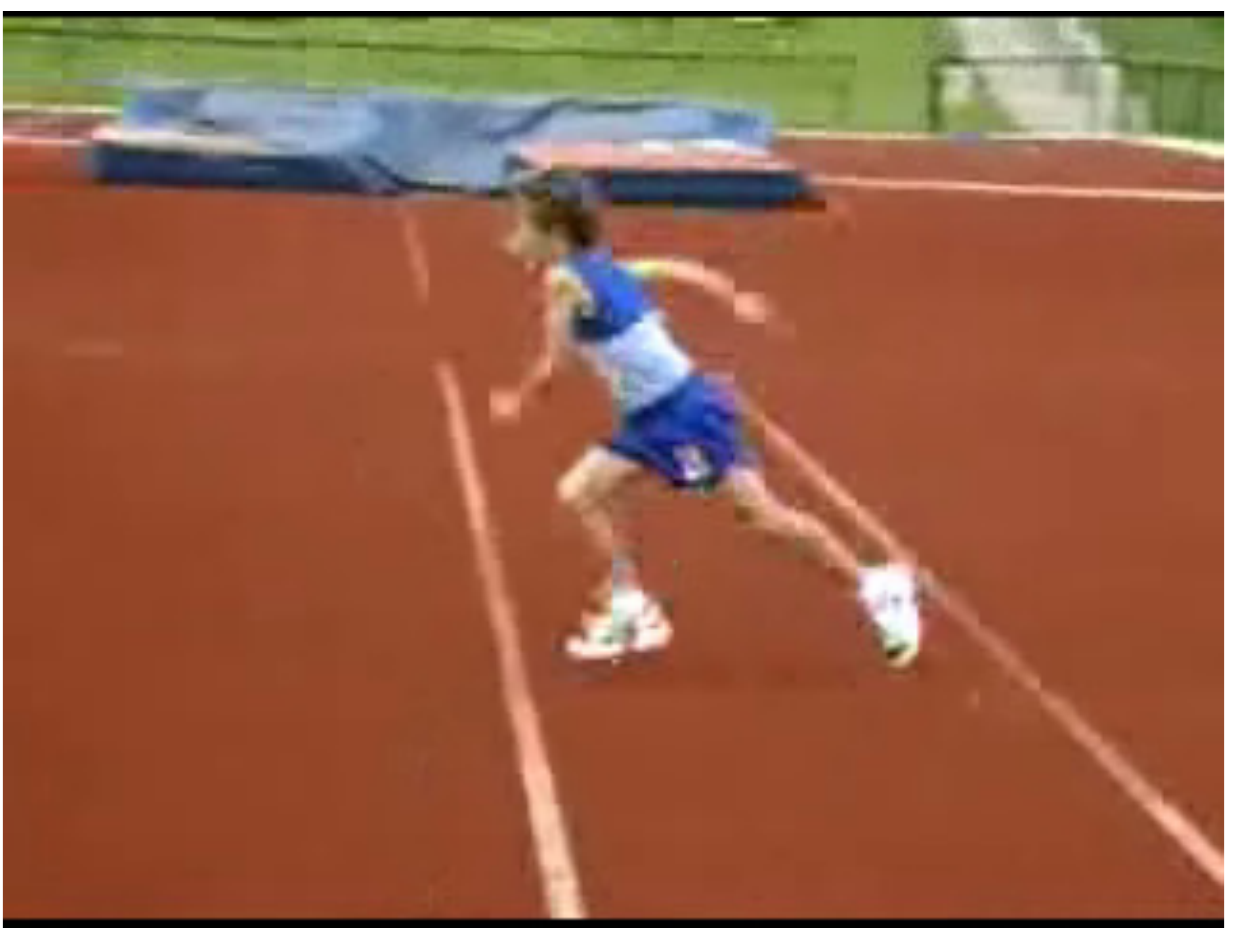}
    \includegraphics[width=0.24\linewidth]{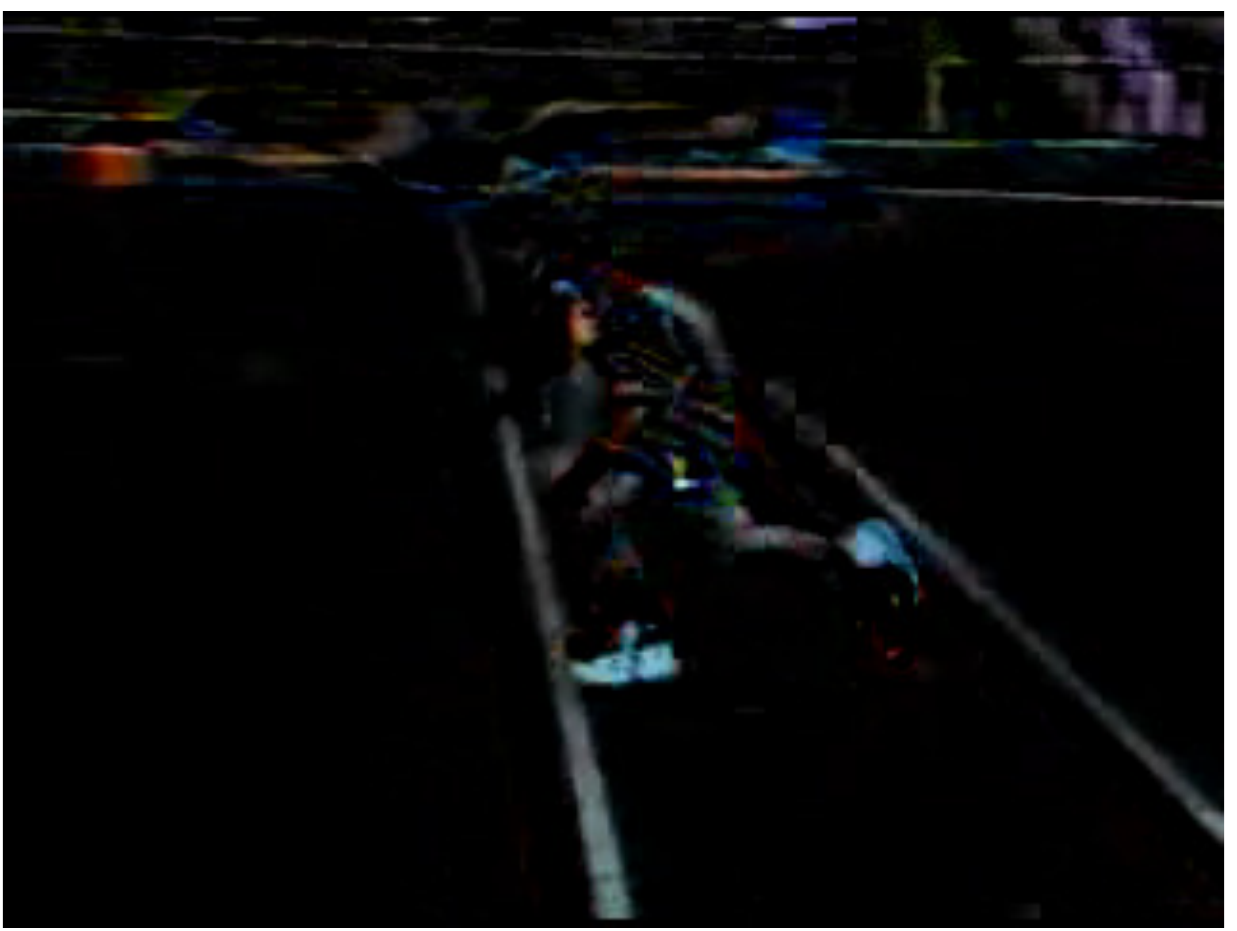}
    \includegraphics[width=0.24\linewidth]{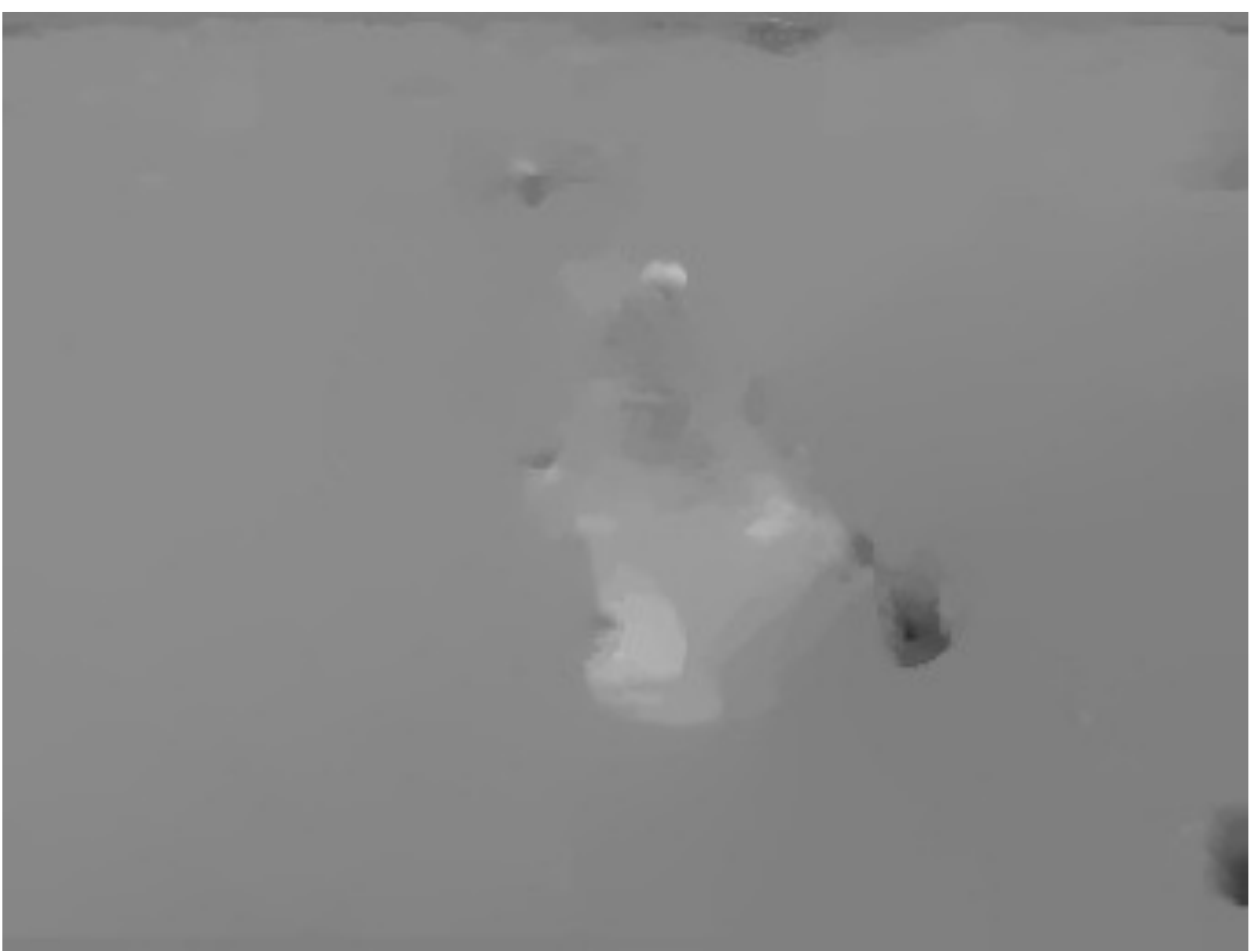}
    \includegraphics[width=0.24\linewidth]{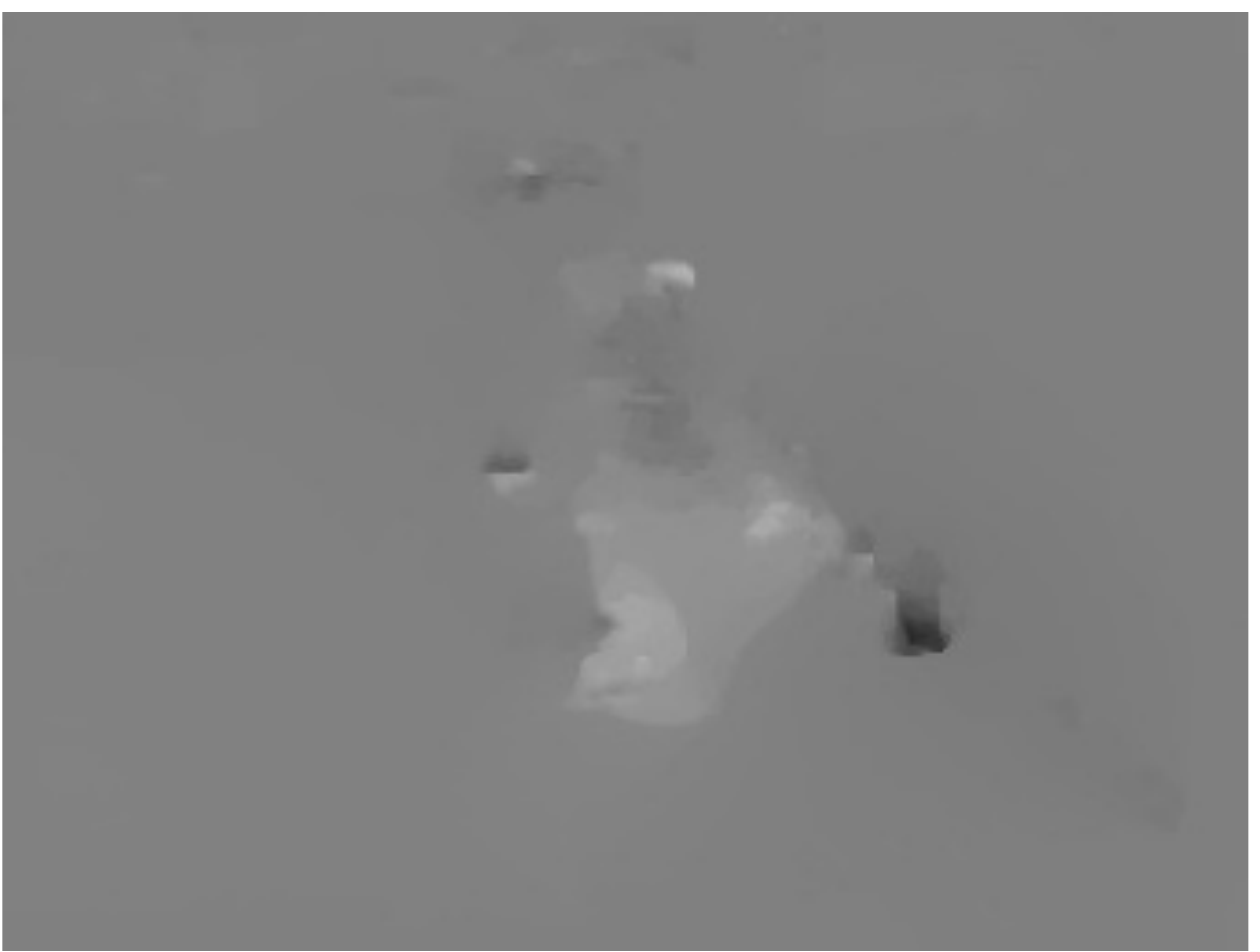} \\
  }
  \vspace{-10pt}
  \caption{Examples of four types of input modality: RGB images, RGB difference, optical flow fields (x,y directions), and warped optical flow fields (x,y directions)}
  \vspace{-15pt}

  \label{fig:ex}
\end{figure}

\textbf{TSN Architectures.}
Our TSN is a general and flexible framework for video-level learning. To demonstrate the generality of our approach, we instantiate TSN with multiple network architectures. Specifically, for {\em ablation studies} on standard action recognition benchmarks, we choose the Inception architecture with Batch Normalization (inception v2)~\cite{IoffeS15} due to its good balance between accuracy and efficiency. Compared with other ConvNet architectures deployed in videos~\cite{SimonyanZ14,TranBFTP15}, this architecture is equipped with better modeling capacity, allowing to demonstrate the improvement of TSN against a strong baseline. In the ActivtyNet Challenge 2016, we investigate more powerful architectures including the Inception V3~\cite{SzegedyVISW16} and ResNet-152~\cite{HeZRS15}, to fully unleash the potential of TSN framework in video classification.

\textbf{TSN Inputs.}
Unlike static images, the additional temporal dimension of videos delivers another important cue for action understanding, namely motion.
In~\cite{SimonyanZ14}, using dense optical flow fields as the source of motion representation is proven to be effective. In this work, we extend this approach in two aspects, namely accuracy and speed. As shown in Figure~\ref{fig:ex}, in addition to the original input modalities of RGB and optical flow~\cite{SimonyanZ14}, we also investigate two other modalities: warped optical flow and RGB differences.

\textit{1) Warped Optical Flow.}
Inspired by the work of improved dense trajectories \cite{WangS13a}, we investigate using warped optical flow fields as the source for motion modeling.
Warped optical flow fields are known to be robust to camera motion and help concentrate on human motion.
We expect this to help to improve the accuracy in motion perception and thus boost the action recognition performance.

\textit{2) RGB Differences.}
Despite the superior recognition accuracy, one issue that impedes the application of two-stream based approaches is the tremendous time cost of optical flow extraction.
To address this problem, we build a motion representation without optical flow.
Inspired by the success of frame volumes~\cite{TranBFTP15} and motion vector~\cite{ZhangWWQW16} in motion modeling, we revisit the simplest cues for apparent motion perception: the stacked differences of RGB pixel intensities between consecutive frames.
Recalling the seminal work on dense optical flow in~\cite{HornB1981}, the partial derivatives of pixel intensities with respect to time play critical roles in computing optical flow.
It is reasonable to hypothesize that the power of optical flow in representing motion could be learned from the simple cues of RGB differences.
This motivates us to investigate using RGB differences as the input of the temporal stream, which greatly saves the time of optical flow extraction.

\textbf{TSN Training.}
As discussed before, existing human annotated datasets for action recognition are limited in terms of sizes.
In practice, training deep ConvNets on these datasets are prone to over-fitting.
To mitigate this issue, we design several strategies to improve the training in the \SEGNET~framework.

\textit{1) Cross Modality Initialization.} 
Pre-training the network parameters on large-scale image recognition datasets, such as ImageNet~\cite{DengDSLL009}, has turned out to be an effective remedy when the target dataset does not have enough training samples~\cite{SimonyanZ14}. As spatial networks take RGB images as inputs, it is natural to exploit models trained on the ImageNet as initialization. For other input modalities such as optical flow and RGB difference, we come up with a cross modality initialization strategy. Specifically, we first discretize optical flow fields into the interval of 0 to 255 by linear transformation. Then, we average the weights of pretrained RGB models across the RGB channels in the first layer and replicate the mean by the channel number of temporal network input.  Finally, the weights of remaining layers of the temporal network are directly copied from the pretrained RGB networks.

\textit{2) Regularization.} 
Batch Normalization \cite{IoffeS15} is able to deal with the problem of covariate shift by estimating the activation mean and variance within each batch to normalize these activation values. This operation speeds up the convergence of training, but also increases the risk of over-fitting in the transfer learning process, due to the biased estimation of mean and variance from a limited number of training samples in target dataset. Therefore, after initialization with pre-trained models, we choose to freeze the mean and variance parameters of all Batch Normalization layers except the first one. As the distribution of optical flow is different from the RGB images, the activation value of first convolution layer will have a distinct distribution and we need to re-estimate the mean and variance accordingly. We call this strategy \textbf{partial BN}.
Meanwhile, we add an extra \textbf{dropout} layer with high dropout ratio (set as 0.8 in experiment) after the global pooling layer to further reduce the effect of over-fitting. 

\textit{3) Data Augmentation.} 
In the original two-stream ConvNets~\cite{SimonyanZ14}, random cropping and horizontal flipping are employed to augment training samples. We exploit two new data augmentation techniques: corner cropping and scale-jittering. In the corner cropping technique, the extracted regions are only selected from the corners or the center of an image to avoid implicitly focusing more on the center area. In the multi-scale cropping technique, we adapt the scale jittering technique \cite{SimonyanZ14a} used in ImageNet classification to action recognition. We present an efficient implementation of scale jittering. We fix the input size as $256 \times 340$, and the width and height of cropped regions are randomly selected from $\{256, 224, 192, 168\}$. Finally, these cropped regions will be resized to $224 \times 224$ for network training. In fact, this implementation not only contains scale jittering, but also involves aspect ratio jittering.

\section{Action Recognition with TSN Models}

With the principled framework of \SEGNET s, there still remains the question of how to use the models learned with this framework to recognize actions in realistic videos. In this section, we describe how to apply action models under two different conditions: trimmed videos and untrimmed videos, and devise a series of techniques in order to improve the robustness of action recognition. 

\subsection{Action Recognition in Trimmed Video}
In trimmed videos, action instances are manually cropped from the long video sequences and thereby action recognition could be simply cast as a classification problem.   
Due to the fact that all snippet-level ConvNets share the model parameters in \SEGNET s, the learned models can perform frame-wise evaluation as normal ConvNets~\cite{SimonyanZ14}.
This also allows us to carry out fair comparison with models learned without the \SEGNET~framework. 
Specifically, we follow the testing scheme of the original two-stream ConvNets \cite{SimonyanZ14}, where we sample $ 25 $ snippets of different modalities. Meanwhile, we crop $ 4 $ corners and $ 1 $ center, and their horizontal flipping from the sampled snippets to evaluate the ConvNets. We use average pooling to aggregate the predictions of different crops and snippets.
For the fusion of predictions from multiple modalities, we take a weighted average of them, where the fusion weights are determined empirically.
It is described in Sec.~\ref{sec:seg} that the segmental consensus function is applied before the Softmax normalization.
To test the models in compliance with their training, we fuse the prediction scores of $ 25 $ frames and different streams before Softmax normalization.

\subsection{Action Recognition in Untrimmed Videos}
\label{sec:untrimmed}

The major obstacle for action recognition in untrimmed videos is the large portion of irrelevant content in the input videos.
Since our action models are trained on trimmed action clips, reusing the technique used for trimmed video, \emph{i.e.}, simply averaging scores from every location in a video, has a high risk of factoring in the unpredictable responses of the models on background contents. 
This makes it necessary to design a specialized method for applying the trained action recognition models to untrimmed videos.
For this purpose, we start by summarizing the following challenges posed by untrimmed videos.

\begin{itemize}
	\item Location issue: an action clip can appear at any temporal location of the video.
	\item Duration issue: the action clip can be either long-lasting or ephemeral.
	\item Background issue: the irrelevant content in a video can have high variations and can possibly occupy a large portion of the whole duration of a video.
\end{itemize}

To deal with these challenges, we develop a detection based method to apply action models to untrimmed videos. First, to cover any location that the action instance can reside, we sample snippets from the input videos in a fixed sampling rate (e.g., 1FPS).
A trained TSN model is then evaluated on these sampled snippets. 
Then, in order to cover the highly varying durations of action clips, a series of temporal sliding windows with different sizes are then applied on the frame scores.
The maximum scores of the classes within a window are used to represent it. 
To alleviate the interference of background contents, windows with the same length are then aggregated with a top-$ \mathcal{K} $ pooling scheme.
The aggregation results from different window sizes then vote for the final prediction of the whole video.

Formally, for a video in length of $ M $ seconds, we will obtain $ M $ snippets $ \{T_1, \ldots, T_M\} $. Applying the TSN model, we will obtain class scores $ \mathcal{F}(T_m) $ for the snippet $ T_m $.
We then build temporal sliding windows with the size of $ l \in \{1, 2, 4, 8, 16\} $. 
The windows will slide through the whole duration of videos, with a stride of $ 0.8 \times l $.
For a window position starting at the $s^{th}$ second, a series of snippets will be covered as $ \{T_{s+1}, \ldots, T_{s+l}\} $, with their class scores $ \{ \mathcal{F}(T_{s+1}), \ldots, \mathcal{F}(T_{s+l}) \}$.
The class scores for this window $ \mathbf{F}^{s,l} $ can be calculated by: 
\begin{equation}
	F^{s,l}_{i} = \max_{p \in \{1, 2 \ldots, l\}}\{f_i^{s+p}\}.
\end{equation}
In this way, for size $ l $ we will obtain $ N^l $ windows, where $ N^l = \lfloor\frac{M}{0.8l}\rfloor $.
Then we apply the aforementioned top-$ \mathcal{K} $ pooling scheme to obtain the consensus $ \mathbf{G}^l $ of from these $N^l$ windows of size $ l $.
Here the parameter $ \mathcal{K} $ is determined as $ \mathcal{K} = \max(15, \lceil N^l/4 \rceil ) $.
This gives us $ 5 $ sets of class scores for window size $ l \in \{1, 2, 4, 8, 16\} $. 
The final score is then calculated as
$ \mathbf{P} = \frac{1}{5} \sum_{l \in \{1,2,4,8,16\}}  \mathbf{G}^l$, which is the average of the five window sizes.
We term this video classification technique as \emph{Multi-scale Temporal Window Integration}, abbreviated as M-TWI.

\section{Experiments}
\label{sec:exp}

In this section, we first introduce the evaluation datasets and implementation details of our approach. 
Then we discuss the practical issues for action recognition with deep learning and our proposed good practices to mitigate them.
After dealing with these issues, we provide detailed analysis of the proposed \SEGNET~framework, to demonstrate the importance of modeling long-term temporal structures.
Finally, we compare the performance of our method with the state of the art on the four action recognition benchmarks. 
We also present the results of our approach in the ActivityNet challenge 2016 and describe the winner solution to this challenge. 
Additionally, we visualize our learned ConvNet models to help qualitatively justify the performance improvement.
If not specifically noted, the experiments in the section are conducted with BN-Inception~\cite{IoffeS15} as the underlying CNN architecture.

\subsection{Datasets}

The datasets adopted to evaluate the performance of \SEGNET~framework are from two types of videos, \emph{i.e.}, trimmed videos and untrimmed videos. Now we describe the details of these datasets.

\textbf{Trimmed Video Datasets.} We conduct experiments on two standard action recognition datasets of trimmed videos, namely HMDB51 \cite{KuehneJGPS11} and UCF101 \cite{Soomro12}. The UCF101 dataset contains $ 101 $ action classes and $13,320$ video clips. We follow the evaluation scheme of the THUMOS13 challenge~\cite{THUMOS13} and adopt the three training/testing splits for evaluation. The HMDB51 dataset is a large collection of realistic videos from various sources, such as movies and web videos. The dataset is composed of $6,766$ video clips from $51$ action categories. Our experiments follow the original evaluation scheme using three training/testing splits and report \textbf{average accuracy} over these splits.

\textbf{Untrimmed Video Datasets.} We conduct experiments of untrimmed video action recognition on two publicly available large-scale datasets. The first is the THUMOS14~\cite{Jiang2014THUMOS14}. It has $ 101 $ classes of human actions. This dataset is composed of training set, validation set, testing set, and background set. We use the training set (UCF101) and validation set ($ 1,010 $ videos) for TSN training and evaluate the learned models on its testing set, which has $ 1,575 $ videos.
The second dataset for untrimmed videos is the ActivityNet~\cite{HeilbronEGN15} dataset. We use its release version 1.2, termed as ActivityNet v1.2. The ActivityNet v1.2 dataset has $ 100 $ classes of human activities. It consists of $ 4,819 $ videos for training, $ 2,383 $ videos for validation, and $ 2,480 $ videos for testing. We follow the standard splits to train and evaluate the our TSN framework.
On both datasets for untrimmed videos, the evaluation metric is \textbf{mean average precision (mAP)} for action recognition.

\subsection{Implementation Details}

We use the mini-batch stochastic gradient descent algorithm to learn the network parameters, where the batch size is set to $ 128 $ and momentum set to $ 0.9 $. We initialize network weights with pre-trained models from ImageNet \cite{DengDSLL009}. We set a smaller learning rate in our experiments. On the dataset of UCF101, for spatial networks, the learning rate is initialized as $0.001$ and decreases to its $\frac{1}{10}$ every $ 1,500 $ iterations. The whole training procedure stops at $ 3,500 $ iterations. For temporal networks, we initialize the learning rate as $0.005$, which reduces to its $\frac{1}{10}$ after $ 12,000 $ and $18,000$ iterations. The maximum iteration is set as $20,000$. To speed up training, we employ a data-parallel strategy with multiple GPUs, implemented with our modified version of Caffe~\cite{JiaSDKLGGD14} and OpenMPI~\footnote{\url{https://github.com/yjxiong/caffe}}. The whole training time on UCF101 is around 0.6 hours for spatial TSNs and 8 hours for temporal TSNs with 8 TITANX GPUs. For other datasets such as HMDB51, THUMOS14, ActivityNet, the learning process is the same with that of UCF101, except that the iteration numbers are adjusted according to the dataset sizes. Concerning data augmentation, we use the techniques of location jittering, horizontal flipping, corner cropping, and scale jittering, as specified in Section \ref{sec:training}. For the extraction of optical flow and warped optical flow, we choose the TVL1 optical flow algorithm \cite{ZachPB07} implemented by OpenCV with CUDA.

\subsection{Effectiveness of The Proposed Practices}

In this section, we focus on investigating the effect of the good practices described in Sec.~\ref{sec:training}, including the training strategies and the input modalities. In this exploration study, we use the two-stream ConvNets with very deep architecture adapted from~\cite{IoffeS15}.

{\bf Different learning strategy.} Compared with the original two-stream ConvNets~\cite{SimonyanZ14}, we propose two new training strategies in Section~\ref{sec:training}, namely cross modality pre-training and partial BN with dropout. Specifically, we compare four settings on the split1 of UCF101: (1) training from scratch; (2) only pre-train spatial stream as in~\cite{SimonyanZ14}; (3) with cross modality pre-training; (4) combination of cross modality pre-training and partial BN with dropout. The results are summarized in Table \ref{tbl:learning}. First, we see that the performance of training from scratch is much worse than that of the original two-stream ConvNets (baseline), which implies carefully designed learning strategy is necessary to reduce the risk of over-fitting, especially for spatial networks. Then, we resort to the pre-training of the spatial stream and cross modality pre-training of the temporal stream to help initialize two-stream ConvNets and it achieves better performance than the baseline. We further utilize the partial BN with dropout to regularize the training procedure, which boosts the recognition performance to $92.0\%$. Therefore, in the remaining experiments, we employ all these good practices for model training.

\begin{table}[t]
  \begin{center}
    \caption{Exploration of different training strategies for two-stream ConvNets on the UCF101 dataset ({\bf split 1}). Here, ``from scratch'' refers to the case we initialize the CNN parameters with Gaussian distribution. Experiments here are conducted \textbf{without} TSN.}
    \label{tbl:learning}
    \vspace{-10pt}
    \begin{tabular}{|l|c|c|c|}
      \hline
      Training setting & Spatial  & Temporal & Two-Stream \\
      \hline 
      Baseline~\cite{SimonyanZ14} & $ 72.7\% $ & $ 81.0\% $ & $  87.0\%  $\\
      \hline
      From Scratch  &  $ 48.7\% $ &  $ 81.7\% $ & $ 82.9\% $ \\
      \hline
      Pre-train Spatial  & $ 84.1\% $ &  $ 81.7\% $ & $ 90.0\% $ \\
      \hline
      + Cross modality pre-training  & $ 84.1\% $ &  $ 86.6\% $ & $ 91.5\% $ \\
      \hline
      + Partial BN with dropout & $ 84.5 \% $ & $ 87.2\% $ & $ 92.0\% $ \\
      \hline
    \end{tabular}
    \vspace{-10pt}
  \end{center}
\end{table}

\begin{table}[]
	\centering
	\caption{Exploration of different combination of modalities with TSN on the UCF101 dataset ({\bf over three splits}). In this table, ``RGB'' refers to the RGB video frame stream. ``Flow'' refers to modality of optical flow input. ``Warp'' refers to the modality of warped optical flow. ``RGB Diff.'' refers to the modality using differences of RGB video frames. The speed for testing is evaluated on a TitanX GPU. In the lower half of the table, we compare ``RGB+RGB Diff.'' with other real-time action recognition methods (FPS $>$ 30).}
  \vspace{-10pt}
	\label{tbl:modality_speed}
	\begin{tabular}{|l|l|l|l|}
		\hline
		Modalities       & TSN & Accuracy & Speed (FPS) \\ \hline
		RGB+Flow         & No   & $92.4\%$ & 14          \\ \hline
		RGB+Flow         & Yes  & $94.9\%$ & 14          \\ \hline
		RGB+Flow+Warp    & Yes  & $95.0\%$ & 5           \\ \hline\hline
		Enhanced MV~\cite{ZhangWWQW16}      & -    & $86.4\%$ & 390         \\ \hline
		Two-Stream 3DNet~\cite{DibaPV2016} & -    & $90.2\%$ & 246         \\ \hline
		RGB+RGB Diff.    & No   & $86.8\%$ & 340         \\ \hline
		RGB+RGB Diff.    & Yes  & $91.0\%$ & 340         \\ \hline
	\end{tabular}
  \vspace{-10pt}
\end{table}

{\bf Different input modalities.} We propose two new types of modalities in Section~\ref{sec:training}: RGB difference and warped optical flow fields. 
We try combining different modalities and report the results in Table~\ref{tbl:modality_speed}. 
These experiments are carried out with all the good practices verified in Table~\ref{tbl:learning}. 
We perform multiple experiments with or without TSN (7 segments) to investigate the performance of different input modalities.
We first observe that RGB and optical flow, which is the basic combination in the two-stream ConvNets also works well with TSN, yielding recognition accuracy of $ 94.9\% $. Then we observe that the warped optical flow  slightly increases the performance ($94.9\%$ to $95.0\%$), but severely slows down the testing speed to only $ 5 $ FPS. So we only use RGB and optical flow to report the final performance.
Another interesting finding is that the simple motion representation of RGB differences, when used together with RGB data under TSN framework, can provide competitive recognition performance ($ 91.0\% $) while running at a very fast speed of $ 340 $FPS. 
It also outperforms other state-of-the-art real-time action recognition methods as shown in Table~\ref{tbl:modality_speed}.
This suggests that ``RGB + RGB Diff.'' can serve well for building real-time action recognition systems with moderate accuracy requirement.

\begin{table*}[t]
  \begin{center}
    \caption{Exploration of different segment numbers $ K $ in \SEGNET s on the UCF101 dataset ({\bf over three splits}) and the ActivityNet dataset (train on the training set and test on the validation set). We use the average consensus function in these experiments.}
    \label{tbl:seg_num}
    \vspace{-10pt}
    \begin{tabular}{|l|ccc|ccc|}
      \hline
      Dataset & \multicolumn{3}{|c|}{UCF101} & \multicolumn{3}{|c|}{ActivityNet} \\
      \hline
      $ K $ & Spatial  & Temporal & Two-Stream (1:1) & Spatial  & Temporal & Two-Stream (1:0.5) \\
      \hline
      1 & $ 85.0\%  $& $ 88.3\%$ & $ 92.4\% $ &  $ 82.0\% $ & $ 61.4\% $ & $ 84.7\% $ \\
      \hline 
      3 & $ 86.5\%  $& $ 89.8\% $ & $ 94.2\% $   & $ 83.6\% $ & $ 70.6\% $ & $ 86.9\% $ \\
      \hline
      5  &  $ 86.7\% $ &  $ 90.1\% $ & $ 94.7\% $  & $ 84.6\% $ &$ 72.9\% $ & $ 87.6\% $ \\
      \hline
      7 & $ 86.4\% $ & $  90.1\% $ & $ \mathbf{94.9\%} $  & $ 84.0\% $ &$ 72.8\% $ & $ 87.8\% $  \\
      \hline
      9 & $ 86.2\% $ & $  89.7\% $ & $ \mathbf{94.9\%} $    & $ 83.7\% $& $ 72.6\% $& $ \mathbf{87.9\%} $\\
      \hline
    \end{tabular}
  \end{center}
  \vspace{-15pt}
\end{table*}

\begin{table*}[t]
  \begin{center}
    \caption{Exploration of different segmental consensus functions for \SEGNET s on the UCF101 dataset ({\bf over three splits}) and the ActivtyNet dataset (train on the training set and test on the validation set). We set segment number $K$ as 7 in these experiments.}
    \label{tbl:segmental}
    \vspace{-10pt}
    \begin{tabular}{|l|ccc|ccc|}
    \hline
     Dataset &\multicolumn{3}{|c|}{UCF101} & \multicolumn{3}{|c|}{ActivityNet}   \\
      \hline
      Consensus Function & Spatial& Temporal  & Two-Stream (1:1) & Spatial   & Temporal  & Two-Stream (1:0.5) \\
      \hline 
      Max Pooling & $ 84.9\%  $& $ 83.5\% $ & $ 92.4\% $ & $81.8\%$& $ 62.0\%$ & $ 85.4\% $ \\
      \hline
      Average Pooling  &  $ 86.4\% $ &  $ 90.1\% $ & $ \mathbf{94.9\%} $ & $84.0\%  $& $ 72.8\% $ & $ 87.8\% $ \\
      \hline
      Weighted Average & $ 86.4\% $ & $ 89.7\% $ & $ 94.8\% $ & $83.1\%$ & $ 70.5\%$ & $ 86.4\% $ \\
      \hline
      \hline
      Top-$\mathcal{K}$ Pooling &  $85.5\%$&  $88.8\%$ & $94.2\%$ & $ 84.7 \% $ & $ 73.6\% $ & $ 88.1\% $ \\
      \hline
      Attention Weighting & $86.1\%$ & $89.1\%$ &  $94.6\%$ & $ 84.1\% $   & $ 71.8\% $ & $ \mathbf{88.2}\% $ \\
      \hline
    \end{tabular}
  \end{center}
  \vspace{-15pt}
\end{table*}

\subsection{Empirical Studies of Temporal Segment Networks}

In this subsection, we focus on studying the effectiveness of \SEGNET~framework. 
As described in Sec~\ref{sec:vds}, the framework has two critical components: the sparse snippet sampling scheme and the segment consensus (aggregation) functions. 
To analyze the \SEGNET~framework in-depth, we evaluate the effect of these two components.
In experiments, we follow the bottom-up order of description. 
That is, we first evaluate the sparse snippet sampling scheme. Then we analyze the effect of different consensus (aggregation) function.
The experiments are performed on the datasets of UCF101 and ActivityNet, to reflect the scenarios of both trimmed and untrimmed video action recognition.
We use RGB images and optical flow fields for input modalities in this exploration. 
Finally, to demonstrate the importance of TSN in long-range modeling, we also compare the performance of TSN with other very deep network architectures on the UCF101 dataset.

{\bf Evaluation on segment number.}
The most crucial parameter governing the sparse snippet sampling scheme in TSN is the number of segments $ K $. When $ K $ equals to $ 1 $, TSN degenerates to the plain two-stream ConvNets. Increasing $ K $ is expected to improve the recognition performance of the learned models. In experiments, we vary the number of $ K $ from $ 1 $ to $ 9 $ and evaluate the recognition performance using the same test approaches. 
The results are summarized in Table~\ref{tbl:seg_num}. We observe that increasing the number of segments will generally lead to better performance. For example, the performance of TSN with 7 segments is better than that of TSN with 3 segments (94.9\% vs. 94.2\%). This improvement implies that using more temporal segments will help to capture richer information to better model temporal structure of the whole video . However, when the segment number $ K $ increases from $ 7 $ to $ 9 $, we observe the performance saturate. Thus, to strike a balance between recognition performance and computational burden, we set $ K=7 $ in the following experiments.

{\bf Evaluation on aggregation function.} In~Eq. (\ref{equ:model}), a segmental consensus function is defined by its aggregation function $ \mathcal{G} $, which could be crucial to the final recognition performance.
Here we evaluate five candidates, including the relatively basic: (1) max pooling, (2) average pooling, (3) weighted average, and the more complex: (4) top-$ \mathcal{K} $ pooling and (5) attention weighting, for the form of $ \mathcal{G} $. 
The experimental results are summarized in Table \ref{tbl:segmental}. On UCF101, which consists of trimmed human action videos, the average aggregation function achieves the best performance, and the weight average and attention weighting obtain quite similar performance. On ActivityNet, the top-$ \mathcal{K} $ and attention weighting aggregation functions achieve comparable performance, which outperforms the basic ones such as average pooling. This fact suggests that on datasets with more complex and diverse temporal structure, the advanced aggregation functions will lead to better recognition accuracies.
In this sense, we default to average pooling for short videos (HMDB51 and UCF101) and top-$ \mathcal{K} $ pooling for complex videos (ActivityNet) in later experiments.

{\bf Comparison of CNN architectures.}
We have conducted previous experiments mostly with the BN-Inception architecture. Here we compare the performance of different network architectures on the UCF101 dataset and the results are summarized in Table \ref{tbl:structure}. 
We use $ K=1 $ in these experiments, which is equivalent to the original two-stream ConvNets. Specifically, we compare the performance of four very deep architectures: BN-Inception~\cite{IoffeS15}, GoogLeNet~ \cite{SzegedyLJSRAEVR14}, VGGNet-16~\cite{SimonyanZ14a}, and ResNet-152~\cite{HeZRS15}. The results of different architectures are directly cited from the corresponding references
Among the compared architectures, the very deep two-stream ConvNets adapted from BN-Inception~\cite{IoffeS15} achieves the best accuracy of $92.4\%$, which is still better than the ResNet-152 by $0.6\%$. This performance improvement may be ascribed to the good practices proposed by our approach.
Furthermore, when trained with TSN ($ K=7 $), the accuracy is boosted to $ 94.9\% $. This clearly justifies the effectiveness of modeling long-range temporal structures with TSN.

\begin{table}[t]
  \begin{center}
    \caption{Exploration of different very deep ConvNet architectures on the UCF101 dataset ({\bf over three splits}). 
      ``BN-Inception+TSN'' refers to the setting where the \SEGNET framework is applied on top of the best performing BN-Inception~\cite{IoffeS15} architecture.}
    \label{tbl:structure}
    \vspace{-10pt}
    \begin{tabular}{|l|c|c|c|}
      \hline
      Training setting & Spatial  & Temporal  & Two-Stream \\
      \hline 
      VGG-M~\cite{ChatfieldSVZ14} (from~\cite{SimonyanZ14}) & $ 73.0\% $ & $ 83.7\% $ & $ 86.9\% $ \\
      \hline
      GoogLeNet~\cite{SzegedyLJSRAEVR14} (from~\cite{WangXW015})  & $ 75.3\% $ &  $ 85.8\% $ & $ 89.3\% $ \\
      \hline
      VGGNet-16~\cite{SimonyanZ14a} (from~\cite{WangXW015}) & $ 78.4\% $ &  $ 87.0\% $ & $ 91.4\% $ \\
      \hline
      ResNet-152~\cite{HeZRS15} (from~\cite{FeichtenhoferPW16}) & $83.4\%$ & $87.2\%$ & $91.8\%$ \\
      \hline
      BN-Inception~\cite{IoffeS15} & $ 85.0\% $  & $ 88.3\% $ & $ 92.4\% $ \\
      \hline
      \hline
      BN-Inception+TSN  & $ 86.4 \%$ & $ 90.1 \%$ & $ \mathbf{94.9\%} $ \\
      \hline
    \end{tabular}
  \end{center}
  \vspace{-10pt}
\end{table}

\begin{table}[t]
  \centering
  \caption{Evaluation on the validation set of ActivityNet challenge 2016 data. ``Bn-Inception w/o TSN'' indicates that we train the models without TSN. ``TSN+X'', indicates that we train TSN with underlying the CNN architecture ``X''. ``TSN-Top3'' refers to the case where the top-$ \mathcal{K} $ aggregation function is used with $ \mathcal{K} $ set to $ 3 $. Here we use the fusion weights of 1:0.5 for RGB and optical flow, respectively.}
  \vspace{-10pt}
  \label{tbl:anet_1-3_val}
  \begin{tabular}{|l|c|c|c|}
    \hline
    \multirow{2}{*}{Settings} & \multicolumn{3}{c|}{mAP on ActivityNet v1.3 Val.}                                                         \\ \cline{2-4} 
    & Spatial                           & Temporal                          & Two Stream                        \\ \hline
    BN-Inception w/o TSN      & \multicolumn{1}{c|}{$76.6\%$} & \multicolumn{1}{c|}{$52.7\%$} & \multicolumn{1}{c|}{$78.9\%$} \\ \hline
    TSN + BN-Inception        & $79.7\%$                      & $63.6\%$                      & $84.7\%$                      \\ \hline
    TSN + Inception V3        & $83.3\%$                      & $64.4\%$                      & $87.7\%$                      \\ \hline
    TSN-Top3 + Inception V3   & $84.5\%$                      & $64.0\%$                      & $88.0\%$                      \\ \hline
    TSN-Ensemble              & $ 85.9\% $        & $ 68.3\% $      & $\mathbf{89.7\%}$ \\ \hline
  \end{tabular}
  \vspace{-10pt}
\end{table}

\begin{table}[h!]
  \begin{center}
    \caption{Winning entries in the untrimmed video classification task of ActivityNet challenge 2016. We present the recognition accuracies in the form of mAP values and top-$ 1 $ accuracies. The entries were ranked by mAP values in the challenge. }
    \label{tbl:anet_challenge}
    \vspace{-10pt}
    \begin{tabular}{|l|c|c|}
      \hline
      Team & ~~mAP~~ & Top1 Accuracy \\
      \hline
      \hline 
      CES (ours) & $ \mathbf{93.23}\% $ & $ \mathbf{88.14}\% $\\
      QCIS &  $ 92.41\% $ & $ 87.79\% $\\
      MSRA & $ 91.94\% $ & $ 86.69\% $ \\
      UTS & $ 87.16\% $ & $ 84.90\% $ \\
      Tokyo Univ. & $86.46\%$ & $80.43\%$ \\ 
      \hline
    \end{tabular}
  \end{center}
  \vspace{-20pt}
\end{table}

\subsection{Comparison with The State of The Art}
\begin{table*}[t!]
  \small
  \centering
  \caption{Comparison of our method based on \SEGNET (TSN) with other state-of-the-art methods on the datasets of HMDB51, UCF101, THUMOS14, and ActivityNet (train on the train+val set and evaluate on the test set). We separately present the results of models learned with TSN (3 segments) and TSN (7 segments).}
  \vspace{-10pt}
  \resizebox{1\linewidth}{!}{
  \begin{tabular}{|lr|lr|lr|lr|}
    \hline
    \multicolumn{2}{|c|}{HMDB51} & \multicolumn{2}{|c|}{UCF101} & \multicolumn{2}{|c|}{THUMOS14} & \multicolumn{2}{|c|}{ActivityNet} \\
    \hline
    \hline

    iDT+FV \cite{WangS13a} & $ 57.2\% $ & iDT+FV \cite{WangS13b} & $ 85.9\% $ & iDT+FV~\cite{WangS13b} & $63.1\%$ & iDT+FV~\cite{WangS13b} & $66.5\%$ \\
    DT+MVSV \cite{CaiWPQ14} & $ 55.9\% $ & DT+MVSV \cite{CaiWPQ14} & $ 83.5\% $ &  object+motion~\cite{Jain2015} & 71.6\% & Depth2Action~\cite{ZhuNS2016} & $78.1\%$ \\
    iDT+HSV \cite{PengWWQ14} & $ 61.1\% $ & iDT+HSV \cite{PengWWQ14} & $ 87.9\% $ & & & & \\
    MoFAP \cite{WangQT15b} & $ 61.7\% $ & MoFAP \cite{WangQT15b} & $ 88.3\% $ & & & & \\ 
    \hline
    \hline
    Two Stream \cite{SimonyanZ14} & $ 59.4\% $ & Two Stream \cite{SimonyanZ14} & $ 88.0\% $ & Two Stream~\cite{SimonyanZ14} & $66.1\%$ & Two Stream~\cite{SimonyanZ14} & $71.9\%$ \\
    VideoDarwin \cite{FernandoGMGT15} & $ 63.7\% $ & C3D (3 nets) \cite{TranBFTP15} & $ 85.2\% $ & EMV+RGB~\cite{ZhangWWQW16} & $61.5\%$ & C3D~\cite{TranBFTP15} & $74.1\%$ \\
    MPR \cite{NiMYY15} & $ 65.5\% $ & Two stream +LSTM \cite{Ng15} & $ 88.6\% $ & & & &  \\
    $\mathrm{F_{ST}CN}$\cite{SunJYS15} & $ 59.1\% $ & $\mathrm{F_{ST}CN}$  \cite{SunJYS15} & $ 88.1\% $ & & & & \\
    
    TDD+FV \cite{WangQT15a} & $ 63.2\% $ & TDD+FV \cite{WangQT15a} & $ 90.3\% $ & & & & \\
    LTC~\cite{varol} & $ 64.8\% $ & LTC~\cite{varol} & $ 91.7\% $ & & & & \\
    KVMF~\cite{ZhuW2016} & $ 63.3\% $ & KVMF~\cite{ZhuW2016} & $ 93.1\% $ & & & & \\
    \hline
    \hline
    TSN (3 seg) & $ 70.7\% $ & TSN (3 seg) & $ 94.2\% $ & TSN (3 seg) & $ 78.8\% $ & TSN (3 seg) & $ 89.0\%$ \\
    TSN (7 seg) & $ \mathbf{71.0\%} $ & TSN (7 seg) & $ \mathbf{94.9\%} $ & TSN (7 seg) & $\mathbf{80.1\%}$  & TSN (7 seg) & $ \mathbf{89.6\%} $\\
    \hline
  \end{tabular}
  }
  \label{tbl:stoa}
  \vspace{-10pt}
\end{table*}

After analyzing the effect of the components in \SEGNET s and coming to a reasonable setting, we now compare our action recognition approach against the state-of-the-art methods on both trimmed videos and untrimmed videos. We conduct experiments on four action recognition datasets. The first two, HMDB51 and UCF101, are composed of trimmed videos. The last two, THUMOS14 and ActivityNet v1.2, consist of untrimmed videos. We expect the experimental results on these datasets would provide a thorough comparison with the existing state-of-the-art methods. In experiments, we use the RGB and optical flow modalities to make fair comparison with previous methods. 

{\bf Trimmed Video Datasets.}
We experiment on two challenging trimmed video datasets: HMDB51 and UCF101. The results are summarized in the left columns of Table~\ref{tbl:stoa}, where we compare our method with both traditional approaches such as improved dense trajectories (iDTs)~\cite{WangS13a}, MoFAP representations \cite{WangQT15b}, and deep learning representations, such as 3D convolutional networks (C3D) \cite{TranBFTP15}, trajectory-pooled deep-convolutional descriptors (TDD) \cite{WangQT15a}, factorized spatio-temporal convolutional networks ($\mathrm{F_{ST}CN}$) \cite{SunJYS15}, long term convolution networks (LTC)~\cite{varol}, and key volume mining framework (KVMF)~\cite{ZhuW2016}.
We present the results of TSN with $ 3 $ and $ 7 $ segments with the average aggregation function. We fuse the prediction scores of RGB and optical flow modalities with equal weights (1:1).
Our best results outperform other methods by $ 5.5\% $ on the HMDB51 dataset, and $ 1.8\% $ on the UCF101 dataset. The superior performance of our method demonstrates the effectiveness of \SEGNET~on trimmed videos and the importance of effective long-term temporal modeling.

{\bf Untrimmed Video Datasets.}
We also compare our approach with other methods on two untrimmed video datasets: THUMOS14 and ActivityNet v1.2. The results are summarized in the right columns of Table~\ref{tbl:stoa}. We compare TSN with the existing methods for untrimmed video action recognition, including improved dense trajectories (iDTs)~\cite{WangS13a}, two-stream ConvNet~\cite{SimonyanZ14}, enhanced motion vectors~\cite{ZhangWWQW16}, 3D convolutional networks~\cite{TranBFTP15}, object+motion~\cite{Jain2015}, and Depth2Action~\cite{ZhuNS2016}. We also present the results of TSN with segment numbers of $ 3 $ and $ 7 $ and the aggregation function in TSN is top-$\mathcal{K}$ pooling. 
Our approach clearly outperforms these compared methods. For example, our TSN (7 seg) is better than the previous best performance by $8.5\%$ on the THUMOS14 dataset and $11.5\%$ on the ActivityNet dataset. This confirms that models learned with TSN also perform well in untrimmed videos, given a reasonable testing scheme, as described in Sec.~\ref{sec:untrimmed}.

\begin{figure*}[t!]
	\centering
	\includegraphics[width=\textwidth]{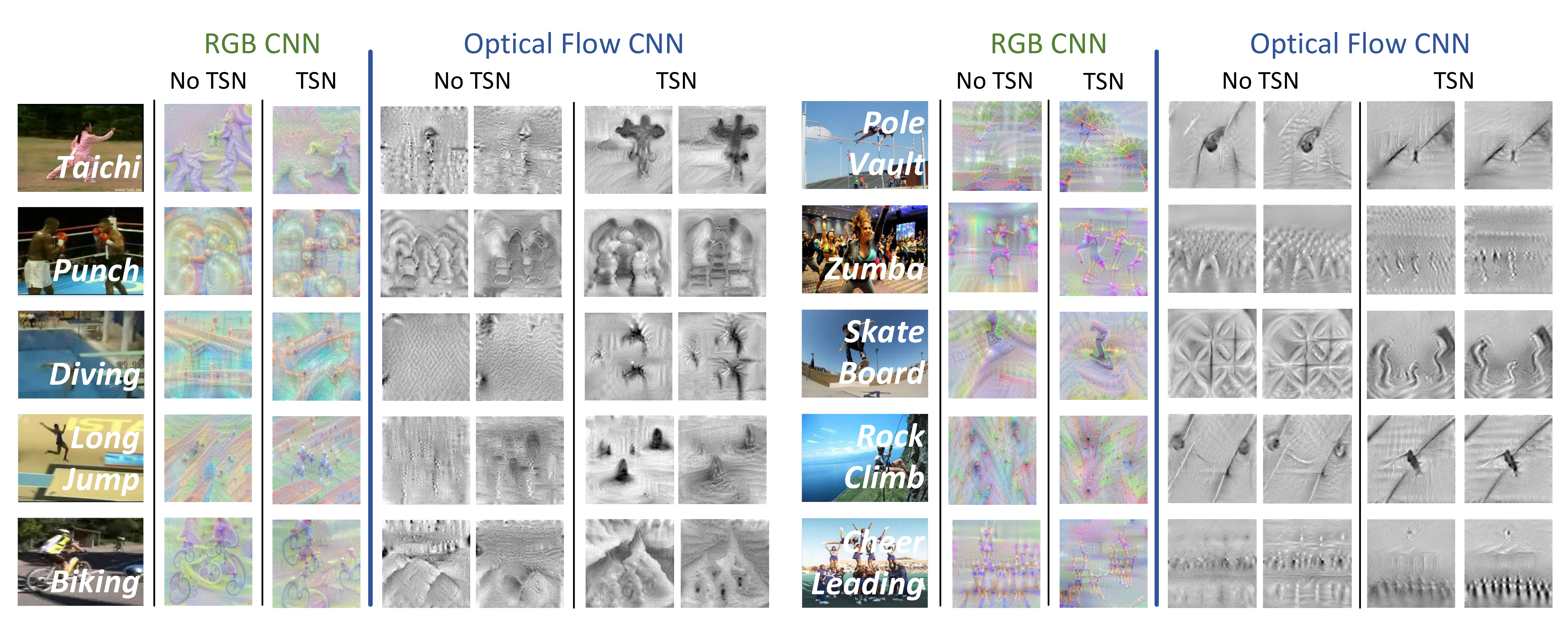}
	\vspace{-25pt}
	\caption{Visualization of ConvNet models for action recognition using DeepDraw~\cite{DeepDraw}. 
		We compare two settings: (1) without \SEGNET~ (No TSN); (3) with \SEGNET ~(TSN).
		For spatial ConvNets, we plot two generated visualization as color images.
		For temporal ConvNets, we plot the flow maps of $ x $ (left) and $ y $ (right) directions in gray-scales.
		Note all these images are generated from purely random pixels.
		\textbf{Left}: classes in UCF101. \textbf{Right}: classes in ActivityNet v1.2.}
	\label{fig:visualization}
	\vspace{-10pt}	
\end{figure*}

\subsection{ActivityNet Challenge 2016}

The power of the temporal segment network framework is further verified in the ActivityNet large scale activity recognition challenge 2016. 
In this challenge we use the videos from the ActivityNet~\cite{HeilbronEGN15} version 1.3 for training and testing.
In Particular, we train TSN models using the trimmed activity instances from the ActivityNet v1.3. 
To test the models, we follow the approach described in Sec.~\ref{sec:untrimmed}.
Understanding that the underlying CNN architecture plays an important role in boosting the performance, we also instantiate TSN with the ultra-deep Inception V3~\cite{SzegedyVISW16} and ResNet~\cite{HeZRS15} architectures.

To evaluate the performance of TSN, we experiment with two settings.
First we train the models on the ``training'' subset of ActivityNet v1.3 and test the recognition accuracy in terms of mean average precision (mAP) on the ``validation'' subset.
In the second setting, we train the models with both ``training'' and ``validation'' subsets and test the recognition accuracy on the ``testing'' subset.
The mAP values on the ``testing'' subset are reported by the publicly available test server of the challenge~\footnote{http://activity-net.org/challenges/2016/evaluation.html}.
The results on validation set are summarized in Table~\ref{tbl:anet_1-3_val}. We observe that TSN significantly boosts the performance over plain two-stream ConvNets (from $78.9\%$ to $84.7\%$). The performance gain is further amplified by using deep CNN architectures such as Inception V3.
Also, the advanced aggregation function such as Top-$\mathcal{K} $ pooling leads to even better performance.
After all, we find that models trained with different aggregation functions (i.e., average pooling, Top-$\mathcal{K}$ pooling, attention weighting) and CNN architectures (i.e., Inception V3, ResNet-152) are complementary when combining into an ensemble, leading to an mAP value of $ 89.7\% $.

{\bf Challenge solution and result.} The results on the testing set are summarized in Table~\ref{tbl:anet_challenge}.
Out entry ``CES'' ranks first among all 24 challenge participants with an mAP of $ 93.23\% $ on the testing set.
The submission is an ensemble of TSN models trained on training and validation data with Inception V3 and ResNet-152 architectures, and the audio models~\cite{ZhuEH16} trained on audio signals of the videos.
For references we also list the results from other participants of this challenge in Table~\ref{tbl:anet_challenge}.
It is worth noting that thanks to the high efficiency of TSN, our models in the challenge can be trained within $ 10 $ hours on a single node with $ 8 $ TitanX GPUs.

\subsection{Model Visualization}

Besides recognition accuracies, we would like to attain further insight into the learned ConvNet models. In this sense, we adopt the DeepDraw~\cite{DeepDraw} toolbox. This tool conducts iterative gradient ascent on input images with only white noises. Thus the output after a number of iterations can be considered as class visualization based solely on class knowledge inside the ConvNet model. The original version of the tool only deals with RGB data. To conduct visualization on optical flow based models, we adapt the tool to work with our temporal ConvNets. As a result, we for the first time visualize interesting class information in action recognition ConvNet models. We randomly pick five classes from the UCF101 dataset,~\emph{Taichi},~\emph{Punch},~\emph{Diving},~\emph{Long Jump}, and~\emph{Biking}, and five classes from the ActivityNet dataset, ~\emph{Poole Vault},~\emph{Zumba},~\emph{Skate Board},~\emph{Rock Climb}, and~\emph{Cheer Leading}. The results are shown in Fig.~\ref{fig:visualization}. For both RGB and optical flow, we visualize the ConvNet models learned with following two settings: (1) training without temporal segment network and (2) training with \SEGNET.

It is also easy to notice that the models, trained with only short-term information such as single frames, tend to mistake the scenery patterns and objects in the videos as significant evidences for action recognition. 
For example, in the class ``Diving'', the single-frame spatial stream ConvNet mainly looks for water and diving platforms, other than the person performing diving. Its temporal stream counterpart, working on optical flow, tends to focus on the motion caused by waves of surface water. 
With long-term temporal modeling introduced by \SEGNET , it becomes obvious that learned models focus more on humans in the videos, and seem to be modeling the long-range structure of the action class. Still consider ``Diving'' as the example, the spatial ConvNet with \SEGNET now generate an image that human is the major visual information. And different poses can be identified in the image, depicting various stages of one diving action. Similar observation would be identified in other action classes such as ``Long Jump'' and ``Skate Board''.
This suggests that models learned with the proposed method may perform better, which is well reflected in our quantitative experiments. 

\section{Conclusions}
\label{sec:con}

In this paper, we presented the Temporal Segment Network (TSN), a video-level framework that aims to model long-range temporal structure. As demonstrated on four action recognition benchmarks and ActivtyNet challenge 2016, this work has brought the state of the art to a new level, while maintaining a reasonable computational cost. This is largely ascribed to the segmental architecture with sparse sampling, as well as a series of good practices that we explored in this work. The former provides an effective and efficient way to capture long-range temporal structure, while the latter makes it possible to train very deep networks on a limited training set without severe overfitting.\\

\bibliographystyle{IEEEtran}
\bibliography{deep}

\end{document}